# MTalk-Bench: Evaluating Speech-to-Speech Models in Multi-Turn Dialogues via Arena-style and Rubrics Protocols

**Yuhao Du**\* **Qianwei Huang**\* **Guo Zhu** **Zhanchen Dai** **Shunian Chen** **Qiming Zhu**
**Le Pan** **Minghao Chen** **Yuhao Zhang** **Li Zhou** **Benyou Wang**[†] **Haizhou Li**
School of Data Science, The Chinese University of Hong Kong, Shenzhen
https://freedomintelligence.github.io/MTalk-Bench/

## Abstract

The rapid advancement of speech-to-speech (S2S) large language models (LLMs) has significantly improved real-time spoken interaction. However, current evaluation frameworks remain inadequate for assessing performance in complex, multi-turn dialogues. To address this, we introduce **MTalk-Bench**, a multi-turn S2S benchmark covering three core dimensions: *Semantic Information*, *Paralinguistic Information*, and *Ambient Sound*. Each dimension includes nine realistic scenarios, along with targeted tasks to assess specific capabilities such as reasoning. Our dual-method evaluation framework combines *Arena-style* evaluation (pairwise comparison) and *Rubrics-based* evaluation (absolute scoring) for relative and absolute assessment. The benchmark includes both model and human outputs, evaluated by human evaluators and LLMs. Experimental results reveal two sets of findings. **Overall performance of S2S LLMs:** (1) models excel at semantic information processing yet underperform on paralinguistic information and ambient sounds perception; (2) Models typically regain coherence by increasing response length, sacrificing efficiency in multi-turn dialogues; (3) modality-aware, task-specific designs outperform brute scaling. **Evaluation framework and reliability:** (1) Arena and Rubrics yield consistent, complementary rankings, but reliable distinctions emerge only when performance gaps are large; (2) LLM-as-a-judge aligns with humans when gaps are clear or criteria explicit, but exhibits position and length biases and is reliable on nonverbal evaluation only with text annotations. These results highlight current limitations in S2S evaluation and the need for more robust, speech-aware assessment frameworks.

## 1 Introduction

End-to-end S2S LLMs represent a major advance in human-computer interaction, enabling natural, direct speech-based communication Jia et al. (2019); Gupta et al. (2024); Communication et al. (2023). However, their rapid progress has outpaced the development of evaluation frameworks, particularly for complex multi-turn dialogues crucial to real-world use. Without holistic context-aware benchmarks, it remains difficult to comprehensively measure progress or diagnose model limitations.

Current evaluation frameworks for speech models are often fragmented, assessing isolated sub-tasks rather than the integrated S2S process. For example, VoiceBench Chen et al. (2024) emphasizes single-turn text-to-speech quality, while ADU-Bench Huang et al. (2024) targets spoken dialogue understanding. Though multi-turn datasets like VoxDialogue Chung et al. (2020) exist, a unified framework for end-to-end S2S evaluation is still lacking. This fragmented approach does not fully reflect the compounded demands of real conversation, including semantic consistency Jurafsky and Martin (2009), paralinguistic cues Schuller and Batliner (2013), and perception of ambient acoustic conditions Barker et al. (2015).

---

\*Equal contribution. ✉ yuhaodu1@link.cuhk.edu.cn, qianweihuang@link.cuhk.edu.cn
[†]Corresponding author. ✉ wangbenyou@link.cuhk.edu.cn

| Benchmarks | Types | | Evaluation Dimensions | | | Input Source | Audio-based Evaluation | Evaluation Method |
|---|---|---|---|---|---|---|---|---|
| | Dialogue | Multi-turn | Semantic | Paralinguistic | Ambient | | | |
| SUPERB Yang et al. (2021) | ✗ | ✗ | ✓ | Emo | ✗ | ✓ | Partial | Obj-Task (WER, PER, ACC) |
| SLUE Shor et al. (2022) | ✗ | ✗ | ✓ | Emo | ✗ | ✓ | ✗ | ASR(WER), Obj-Task (F1 Score) |
| LeBenchmark Evain et al. (2021) | ✗ | ✗ | ✓ | Emo | ✗ | ✓ | Partial | ASR (WER, CER), Obj-Task (BLEU, CCC) |
| SpokenWOZ Si et al. (2024) | ✓ | ✓ | ✓ | ✗ | ✗ | ✓ | ✗ | ASR (WER) + Obj-Task (BLEU) |
| VoxDialogue Cheng et al. (2025) | ✓ | ✓ | ✓ | ✓ | ✓ | ✓ | ✗ | ASR, Text-Metric, LLM Eval |
| AF-Dialogue Kong et al. (2024) | ✓ | ✓ | ✓ | ✗ | ✓ | ✓ | ✗ | ASR, Obj-Task, Text-Metric, LLM-Eval, Human-Eval |
| VoiceBench Chen et al. (2024) | ✓ | ✗ | ✓ | Emo, Vol, Spd | ✗ | Partial | Partial | ASR, Obj-Task (Accuracy, safety rate), LLM-Eval |
| SD-EVAL Ao et al. (2025) | ✓ | ✗ | ✓ | Emo | ✓ | ✓ | ✗ | ASR, Text Metric LLMEval, Human Eval |
| AirBench Yang et al. (2024) | ✓ | ✗ | ✓ | Emo | ✗ | ✓ | ✗ | Obj-Task (Accuracy, correctness rate), LLM-Eval |
| S2S-Arena Jiang et al. (2025) | ✓ | ✗ | ✓ | ✓ | ✗ | Partial | ✗ | Arena, Human Eval |
| **MTalk-Bench (Ours)** | ✓ | ✓ | ✓ | ✓ | ✓ | ✓ | ✓ | Arena, Rubric-based, Human Eval, LLM Eval |

Table 1: Comparison of benchmarks on spoken dialogue with new evaluation dimensions and structure, with an additional column for Human Speech Input (✓: human-recorded, ✗: TTS-generated, Partial: mixed).

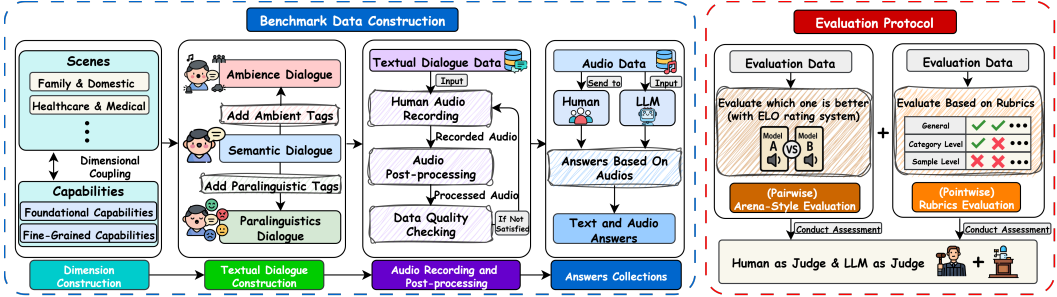

Figure 1: The Overview of MTalk-Bench.

To address this gap, we propose **MTalk-Bench**, the first benchmark designed for holistic evaluation of S2S LLMs in multi-turn settings. It targets three key dimensions of spoken interaction: **Semantic Information**, **Paralinguistic Information**, and **Ambient Sound**. Our dual-method evaluation framework combines **Arena**, for relative model comparison via pairwise voting Zheng et al. (2023), and **Rubrics**, for absolute scoring based on detailed criteria Hashemi et al. (2024). This approach aims to assess both model and human performance, with evaluation conducted by both LLMs and human evaluators. The overview of MTalk-Bench is shown in Figure 1.

Our primary contributions are: (1) We design and present **MTalk-Bench**, the first benchmark for the holistic evaluation of multi-turn S2S dialogue, structured around semantic, paralinguistic, and acoustic dimensions. (2) We propose a novel **Dual-Method Evaluation Framework**, which combines relative pairwise comparisons (*Arena*) with absolute fine-grained scoring (*Rubrics*) to enable comprehensive model assessment. (3) We conduct a **robust analysis** of our evaluation results, revealing significant discrepancies between human and LLM evaluators and highlighting current challenges for a reliable assessment of S2S models.

## 2 Related Works

### 2.1 Speech-to-Speech Models

The recent rapid advances in LLMs Radford et al. (2019); OpenAI et al. (2024) and dialogue-specialized architectures such as DialoGPT Zhang et al. (2020) have significantly improved conversational fluency. Traditional spoken dialogue systems typically adopt cascaded ASR–LLM–TTS pipelines, which incur latency and often lose paralinguistic details. Emerging E2E S2S models, including Translatotron Jia et al. (2019), GLM-4-Voice Zeng et al. (2024), and Qwen2.5-Omni Li et al. (2025), directly map input speech to output speech, enabling richer prosodic control, more faithful emotional preservation, and lower inference latency.

## 2.2 Speech-to-Speech Benchmarks

Despite growing interest in S2S models, most benchmarks remain limited. As summarized in Table 1, existing work typically focuses on single-turn tasks (e.g., SUPERB Yang et al. (2021), SLUE Shor et al. (2022), VoiceBench Chen et al. (2024)), or isolates specific aspects such as detection (e.g., LeBenchmark Evain et al. (2021), SD-EVAL Ao et al. (2025)), while neglecting ambient sound and multi-turn dynamics. Moreover, many solely use text-based scoring or partial audio inputs for evaluation, limiting validity. **MTalk-Bench** addresses these gaps by evaluating semantic, paralinguistic, and ambient sound understanding in natural multi-turn dialogues, using fully audio-grounded inputs and both human and LLM assessments via *arena* and *rubrics*. This positions MTalk-Bench as the first benchmark to support a holistic, speech-native evaluation of S2S models in real-world conversational settings.

## 3 Framework of MTalk-Bench

MTalk-Bench adopts a dual-method evaluation framework, encompassing both comprehensive *scenarios* (i.e., *where* to evaluate) and hierarchical *capabilities* (i.e., *what* to evaluate), detailed in §3.1 and §3.2. The mapping between the two is in §3.3.

### 3.1 User-Centric Evaluation *Scenarios*

**Candidate Scenarios**  Our evaluation scenarios establish realistic contexts that complement our capability taxonomy, defining *what* to evaluate. Initially, we curate *twenty* candidate scenarios derived from extensive literature in communication studies, human-computer interaction (HCI), and linguistics Kuniavsky (2003); Gumperz (1982); Clark (1996); Schegloff (2007). The complete scenario list is detailed in Appendix A.1.

**Scenario Selection via User Voting**  To ground our benchmark in authentic, frequent communication scenarios, we employ a pairwise comparison methodology. Specifically, participants are instructed to select the scenario from each randomly presented scenario pair that they believe is more representative or frequent in typical human-to-human communication contexts. Detailed instructions and the complete survey methodology are available in Appendix A.1. Based on selected scenarios ranking, we choose the *top nine* scenarios. These scenarios provide diverse and representative communicative contexts for assessing the full range of capabilities defined in our taxonomy.

### 3.2 A Hierarchical Taxonomy for *Capabilities*

As shown in Figure 2, MTalk-Bench assesses model competence through a two-tier hierarchical taxonomy, which defines *what* to evaluate. It centers on three core dimensions, based on which, **Tier 1** defines nine *foundational capabilities*. **Tier 2** further breaks these down into fine-grained capabilities, empirically derived from each Tier 1 capability.

**Foundational Capabilities**  Derived from the three dimensions of spoken dialogue, **Tier 1** corresponds to the middle ring of the capability taxonomy. It comprises nine high-level capabilities derived from seminal research in communication science, which are organized into the following groups:

- **Semantic Information**: Understanding & Memory, Reasoning & Execution, Interaction Strategy, Security Assessment, Pragmatic & Culture.
- **Paralinguistic Information**: Paralinguistic Comprehension, Paralinguistic Generation.
- **Ambient Sound**: Ambient Sound Perception, Multi-party Interaction.

**Fine-Grained Capabilities**  Aligned with the foundational dimensions, **Tier 2** corresponds to the outer ring of the capability taxonomy and consists of specific, measurable capabilities. Following a user-centric methodology similar to §3.1, we combine literature review and large-scale pairwise comparisons to identify the most representative capabilities across dimensions.

We rank capabilities using pairwise preference modeling based on the Bradley–Terry method Bradley and Terry (1952); David (1988), ensuring both empirical representativeness and theoretical coverage.

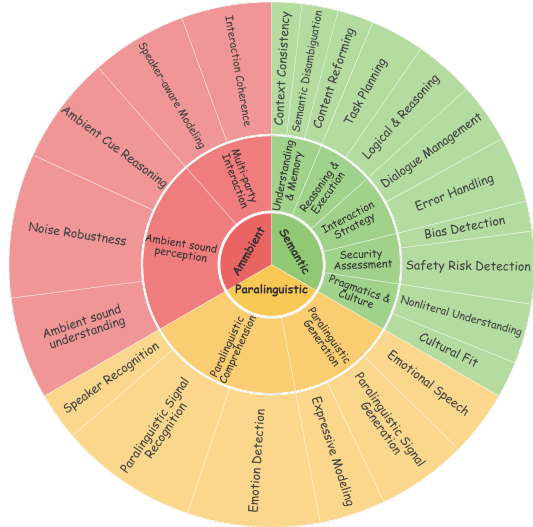

Figure 2: The capability taxonomy of MTalk-Bench

By selecting the highest-ranked capabilities from each foundational dimension, we ensure the empirical validity and theoretical balance of our benchmark. A detailed categorization of these capabilities is provided in Appendix A.2.

### 3.3 *Scenario*-to-*Capability* Mapping

To ensure that the constructed benchmark tasks reflect realistic communicative demands, each selected scenario is mapped to certain capabilities in our taxonomy. This process begins with an analysis of the scenario's real-world requirements across the three dimensions. For each dimension, the most relevant Tier 1 foundational capabilities are identified based on their importance in performing effectively within that scenario. These selections are informed by literature in communication science, domain-specific expertise, and the practical demands observed in authentic contexts.

As an example, consider the *Health and Medical Communication* scenario:

- **Semantic Information:** Effective medical communication requires conveying complex health information clearly and accurately, which is critical for patient understanding and decision-making (Street Jr et al., 2009). This aligns most closely with the *Reasoning & Execution* capability, which supports precise and logical delivery of medical advice.

- **Paralinguistic Information:** Building patient trust depends on recognizing and responding to emotional cues in speech—such as anxiety, uncertainty, or discomfort (Roter et al., 1988). This is best captured by the *Paralinguistic Comprehension* capability, enabling models to interpret subtle prosodic signals that convey affect and intent.

- **Ambient Sound:** Medical environments are often acoustically complex, containing both irrelevant noise and critical auditory cues such as alarms or monitor beeps (Stowell et al., 2015). The *Ambient Sound Understanding* capability ensures that models can filter noise, detect important environmental cues, and maintain robust speech comprehension.

This structured mapping procedure ensures that scenarios are evaluated through a realistic, multi-faceted lens, directly linking communicative context to measurable model capabilities. The complete scenario–capability mapping table is provided in Appendix A.3.

## 4 Benchmark Construction

To ensure authenticity and quality, our evaluation data is constructed through a multi-stage pipeline as illustrated in Figure 3. The process begins with the construction of textual multi-turn dialogues, followed by human audio recording and systematic post-processing. Each instance undergoes multi-round quality assurance to yield the final validated audio dataset.

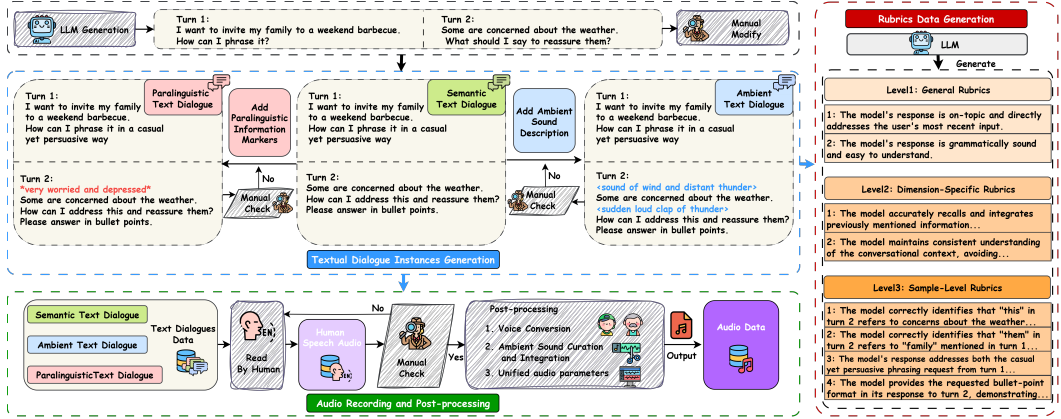

Figure 3: Overall data construction pipeline for MTalk-Bench, from initial textual dialogue generation to rubric creation.

## 4.1 Data Construction

The MTalk-Bench dataset is constructed through a multi-stage pipeline, beginning with high-quality textual multi-turn dialogues for **Semantic Information**. These dialogues are then augmented with annotations for **Paralinguistic Information** and **Ambient Sound**, and finally converted to audio through human recording, synthesis, and mixing. This process ensures both realism and high data quality.

**Constructing Raw Textual Dialogue**   To construct a dataset for evaluating Semantic Information, we first generate over 1,500 raw textual multi-turn dialogue candidates using a hybrid LLM-human pipeline. These candidates are then manually screened for logical coherence, naturalness, and testability. Approximately 19% are retained as high-quality Type I dialogues. Details of the construction and validation process can be found in Appendix B.1.

**Augmenting Dialogue with Paralinguistic and Ambient Sound Tags**   For Paralinguistic and Ambient Sound capabilities, validated Type I dialogues are modified by two annotators to embed expressive cues for Type II and relevant sound descriptions for Type III. A third annotator reviews and resolves discrepancies. Low-quality samples are revised or discarded. Details about annotation are provided in Appendix B.3.

**Audio Recording and Post-processing**   Text dialogues are converted to audio via a structured pipeline. Native English speakers from MTurk record Type I data in a neutral tone and Type II data with emotion guided by annotations. The Seed-VC model Liu (2024) is used to synthesize child or elderly voices. For Type III, real ambient sounds from Freesound Font et al. (2013) and FSD50K Fonseca et al. (2022) are mixed with speech to simulate realistic environments. Futher details about the recording and audio processing are provided in Appendix B.3.

## 4.2 Data Quality Checking

All MTurk audio samples underwent manual review for **semantic accuracy**, referring to adherence to the script, and **paralinguistic fidelity**, referring to the clarity of intended emotions and styles Buhrmester et al. (2011); Scherer (2003). The evaluation proceeded in three rounds. In Round 1, out of 270 samples, 75 were rejected, with 33 rejected for script deviations and 42 for insufficient paralinguistic expression. In Round 2, 75 samples were reviewed and 33 were rejected, including 3 for script deviations and 30 for paralinguistic deficiencies. In Round 3, 33 samples were reviewed and 12 rejected, all for paralinguistic issues. The final batch met all evaluation criteria.

# 5 Evaluation Protocol

Evaluating S2S LLMs requires a comprehensive framework that integrates both relative and absolute perspectives. To achieve this, our protocol employs two complementary methodologies: (1) **Pairwise Arena** (§5.1), which uses head-to-head comparisons and Elo ratings to determine holistic quality based on user preference. (2) **Pointwise Rubrics** (§5.2), which provide absolute, fine-grained scores based on structured criteria for diagnostic analysis. This dual-method approach ensures a robust and interpretable evaluation across all tested dimensions. Complete evaluation protocol design are provided in Appendix C.

## 5.1 Pairwise Arena-Style Evaluation

In the Arena-style protocol, human evaluators perform blind, head-to-head comparisons of model outputs. After reviewing a detailed task guidance, evaluators are presented with the specific inputs for each evaluation: the tested capability, the user's audio input, and two anonymized model responses. Evaluators select the better response based on the target capability and briefly explain their choice.

To ensure robust ranking, the Arena-style uses a dynamic pairing strategy that matches models with similar Elo scores. This improves statistical efficiency and provides high-resolution differentiation between competitive systems.

### 5.1.1 Elo Rating

We use the Elo rating system to quantify and rank model performance in the Pairwise Arena. Each model is initialized with a rating of 1000. Following each pairwise comparison between model A and model B, the rating for model A, $R'_A$, is updated as:

$$R'_A = R_A + K(S_A - E_A) \qquad (1)$$

where $R_A$ is the current rating and $S_A$ is the binary match outcome (1 for a win, 0 for a loss). We set the elasticity coefficient $K = 4$ to ensure stability and minimize the influence of noisy judgments. The expected score for model A, $E_A$, is then calculated as:

$$E_A = \frac{1}{1 + 10^{(R_B - R_A)/400}} \qquad (2)$$

This process creates a dynamic ranking of all models based on their cumulative performance in head-to-head comparisons. More details of the Elo rating system are provided in Appendix C.1.2.

## 5.2 Pointwise Rubric-based Evaluation

While the Arena provides a holistic, relative ranking of models, our Hierarchical Rubrics framework offers a complementary, absolute assessment. This pointwise method evaluates each model response in isolation, scoring it against fine-grained criteria to enable diagnostic analysis. This approach allows for an interpretable breakdown of a model's specific strengths and weaknesses. In this setup, each response is scored against a set of 7 to 9 binary rubrics, receiving a score of 1 if a criterion is met and 0 otherwise.

### 5.2.1 Rubrics Design

Our rubric system is designed with a three-level hierarchy to ensure comprehensive and dimension-specific assessment Suskie (2018):

- **Level 1: General Rubrics** — These universal criteria apply to all responses (e.g., grammatical correctness, relevance to the input).
- **Level 2: Dimension-Specific Rubrics** — These criteria are tailored to the three core dimensions of our benchmark (e.g., context consistency for *Semantic*, emotional clarity for *Paralinguistic*, and background sound awareness for *Ambient Sound*).
- **Level 3: Sample-Specific Rubrics** — These are fine-grained, instance-specific criteria generated by an LLM, based on the unique context of the dialogue and the target capability Hashemi et al. (2024).

To ensure quality, all LLM-generated rubrics (Level 3) are manually reviewed and refined by trained annotators. This human-in-the-loop process combines the scalability of LLMs with the reliability of expert oversight. Detailed rubric annotation is provided in Appendix C.2.2.

### 5.2.2 Rubrics Score

Performance under the Hierarchical Rubrics framework is quantified using an average score. For each of the $M$ test cases, a model's response is scored against $N$ binary criteria ($s_j \in \{0, 1\}$). The score for a single case, $S_{\text{case}}$, is the mean of these criteria scores:

$$S_{\text{case}} = \frac{1}{N} \sum_{j=1}^{N} s_j \tag{3}$$

The model's final score, $\bar{S}_{\text{model}}$, is the average of these scores across all $M$ test cases:

$$\bar{S}_{\text{model}} = \frac{1}{M} \sum_{k=1}^{M} S_{\text{case},k} \tag{4}$$

This score provides an absolute and interpretable measure of a model's capabilities, enabling a clear diagnostic analysis of its strengths and weaknesses. For a more intuitive display, all rubric scores in Table 2 and Appendix E have been multiplied by 100.

| S2S Models | Human | | | | GPT-4o Realtime | | | | Gemini-2.5-pro | | | | Qwen-Omni-Turbo | | | |
|---|---|---|---|---|---|---|---|---|---|---|---|---|---|---|---|---|
| | Sem. | Para. | Ambi. | Ovrl.↑ | Sem. | Para. | Ambi. | Ovrl. | Sem. | Para. | Ambi. | Ovrl. | Sem. | Para. | Ambi. | Ovrl. |
| *Arena-style Evaluation* | | | | | | | | | | | | | | | | |
| *Closed-source Models* | | | | | | | | | | | | | | | | |
| Doubao | 1023 | 1038 | 1049 | 1037 | 1008 | 1029 | 1020 | 1019 | 1012 | 1027 | 1005 | 1015 | 985 | 1017 | 997 | 1000 |
| Qwen-Omni-Turbo | 1007 | 1044 | **1051** | 1034 | 994 | 1020 | 986 | 1000 | 1002 | 982 | 983 | 989 | 1012 | 1007 | 1009 | 1009 |
| GPT-4o Realtime | **1041** | 1029 | 1020 | 1030 | **1036** | 1005 | **1032** | **1024** | 1023 | 1022 | **1028** | **1025** | 1011 | 984 | 1017 | 1004 |
| *Open-source Models* | | | | | | | | | | | | | | | | |
| Step-Audio-Chat | **1058** | **1054** | 1034 | **1049** | 1027 | 1011 | 1009 | 1016 | **1025** | 994 | 1013 | 1010 | 1007 | **1041** | **1050** | **1033** |
| GLM-4-Voice | 1010 | 1011 | 982 | 1001 | 1009 | 1004 | 1015 | 1010 | 1017 | 1023 | 1009 | 1016 | 1018 | 1001 | 1010 | 1010 |
| VITA-Audio-Plus-Vanilla | 1004 | 973 | 1001 | 993 | 1025 | 1004 | 1029 | 1019 | 995 | 999 | 1004 | 999 | **1035** | 994 | 1013 | 1014 |
| MiniCPM-o 2.6 | 989 | 980 | 979 | 983 | 973 | 987 | 986 | 982 | 972 | 984 | 997 | 984 | 992 | 1008 | 994 | 998 |
| Kimi-Audio | 970 | 978 | 972 | 973 | 984 | **1048** | 997 | 1010 | 996 | **1032** | 1025 | 1018 | 990 | 999 | 976 | 988 |
| Moshi | 950 | 970 | 947 | 956 | 987 | 965 | 991 | 981 | 971 | 969 | 969 | 970 | 985 | 984 | 984 | 984 |
| AnyGPT | 942 | 941 | 931 | 938 | 956 | 924 | 960 | 947 | 977 | 980 | 990 | 982 | 1001 | 984 | 962 | 982 |
| Human | 1005 | 981 | 1034 | 1006 | 981 | 970 | 968 | 973 | 998 | 983 | 970 | 983 | 945 | 972 | 960 | 959 |
| *Rubric-based Evaluation* | | | | | | | | | | | | | | | | |
| *Closed-source Models* | | | | | | | | | | | | | | | | |
| GPT-4o Realtime | **88.59** | 73.75 | **69.73** | **77.38** | **88.06** | **76.01** | **67.80** | **77.33** | **73.76** | **64.34** | **70.31** | **69.50** | **84.85** | 80.33 | **86.51** | **84.01** |
| Doubao | 82.54 | 77.06 | 60.42 | 73.69 | 81.97 | 73.61 | 60.50 | 72.06 | 70.02 | 60.31 | 60.63 | 63.69 | 82.73 | 80.28 | 82.07 | 81.71 |
| Qwen-Omni-Turbo | 76.08 | **78.83** | 60.05 | 71.82 | 74.88 | 66.29 | 60.75 | 67.34 | 54.35 | 47.59 | 56.73 | 52.91 | 82.83 | **80.71** | 84.48 | 82.72 |
| *Open-source Models* | | | | | | | | | | | | | | | | |
| Step-Audio-Chat | 85.50 | 70.20 | 59.55 | 71.86 | 79.23 | 67.80 | 59.87 | 69.01 | 62.81 | 55.64 | 56.23 | 58.25 | 81.39 | 79.53 | 84.04 | 81.73 |
| GLM-4-Voice | 72.41 | 75.45 | 64.33 | 70.81 | 79.10 | 74.87 | 63.02 | 72.35 | 57.09 | 57.43 | 57.48 | 57.33 | 82.04 | 80.46 | 82.94 | 81.86 |
| VITA-Audio-Plus-Vanilla | 79.56 | 67.50 | 59.22 | 68.70 | 79.23 | 72.73 | 63.77 | 71.94 | 59.45 | 55.89 | 57.11 | 57.50 | 80.81 | 80.54 | 85.20 | 82.20 |
| MiniCPM-o 2.6 | 63.15 | 57.11 | 46.04 | 55.74 | 71.64 | 70.45 | 58.11 | 66.75 | 49.63 | 51.07 | 49.69 | 50.13 | 75.28 | 75.67 | 78.51 | 76.50 |
| Kimi-Audio | 65.56 | 50.10 | 50.00 | 55.17 | 74.86 | 75.63 | 61.73 | 70.66 | 55.33 | 60.62 | 58.63 | 58.17 | 78.59 | 79.49 | 76.01 | 77.98 |
| Moshi | 35.05 | 46.58 | 29.58 | 37.26 | 46.39 | 47.15 | 37.74 | 43.76 | 29.73 | 32.15 | 28.68 | 30.18 | 61.62 | 58.99 | 60.93 | 60.55 |
| AnyGPT | 28.57 | 37.39 | 32.47 | 32.76 | 56.38 | 64.90 | 43.90 | 55.05 | 15.30 | 11.94 | 7.04 | 11.44 | 37.20 | 38.87 | 44.03 | 40.06 |
| Human | 63.90 | 61.49 | 63.55 | 62.97 | 71.14 | 66.67 | 63.02 | 66.96 | 45.02 | 50.06 | 58.36 | 51.13 | 66.67 | 71.58 | 71.82 | 69.96 |

Table 2: Combined Evaluation Results: Arena-style Elo scores are rounded to the nearest integer. Rows are sorted by the Human Overall (Ovrl.↑) scores in descending order. ↑ indicates rows are ranked by Human Overall scores (high to low). For vote counts, refer to the Appendix E.

## 6 Experiments

Our experiments are designed to comprehensively evaluate S2S LLMs using the dual-method protocol described in §5.

### 6.1 Evaluation Setup

**Benchmarked Models and Baseline** We evaluate a diverse set of S2S LLMs, including leading closed-source systems (GPT-4o Realtime, Doubao, MindGPT-4o-Audio) and prominent open-source models (e.g., GLM-4-Voice-9B, Qwen-Omni-Turbo, Kimi-Audio, VITA-Audio-Plus-Vanilla, and others). To establish a performance baseline, we also include human-generated responses in the evaluation. All model outputs are generated with uniform hyperparameters (*temperature*=0.5, *top_p*=0.95). For models lacking public APIs, we use a standardized microphone capture method. A detailed description of all models, generation configurations, and the computing infrastructure is provided in Appendix D.1.

## 6.2 Human or LLM Evaluators

### 6.2.1 Human Evaluator

Human evaluators, recruited via a public link and the MTurk platform, assesse the **raw audio** outputs from all models. Details on the recruitment and quality control protocols are in Appendix D.3.1. After quality filtering, the following evaluations are retained for analysis:

- **Arena-style Evaluation:** From an initial 1,912 comparisons (213 annotators), we retain **1,602** high-quality evaluations, distributed across Semantic (584), Paralinguistic (555), and Ambient (463) dimensions.

- **Rubrics Evaluation:** From an initial 2,160 assessments (112 annotators), we retain **1,599** valid evaluations, covering Semantic (537), Paralinguistic (537), and Ambient dimensions (525).

### 6.2.2 LLM-as-a-Judge

To explore the feasibility of automated evaluation, we deploy an LLM-as-a-Judge. We used a dual-modality approach to specifically test its understanding of non-verbal cues, assessing S2S LLM performance under the following modalities:

- **Raw Audio:** The LLM directly evaluates the audio files. For this modality, we perform Rubric-based scoring on all 270 samples for all evaluated models and collect approximately 500 pairwise Arena-style comparisons per dimension.

- **Transcribed Text:** The LLM evaluate ASR transcripts enriched with annotations for paralinguistic or ambient sound information (e.g., `[laughs]`). For this modality, we perform Rubric-based scoring on all 270 samples for all evaluated models and collect approximately 300 pairwise Arena-style comparisons per dimension.

Detailed protocols for the LLM-as-a-Judge experiment, along with a comparative analysis of the raw audio and transcribed text results, are presented in Appendix D.3.2.

| | Human | GPT-4o Realtime | Gemini-2.5-pro | Qwen-Omni-Turbo |
|---|---|---|---|---|
| TPR (%) | 49.12 [46.35–51.89] | 51.45 [49.58–53.31] | 52.37 [50.51–54.22] | 48.52 [46.63–50.42] |
| BPR (%) | 50.88 [48.11–53.65] | 48.55 [46.69–50.42] | 47.63 [45.78–49.49] | 51.48 [49.58–53.37] |
| $\Delta$Position Bias | -1.76 | 2.89 | **4.74**[**] | -2.96 |
| LPR (%) | 56.5 [53.6–59.3] | 57.1 [55.2–59.0] | 52.0 [50.1–53.9] | 58.6 [56.7–60.6] |
| SPR (%) | 43.5 [40.7–46.4] | 42.9 [41.0–44.8] | 48.0 [46.1–49.9] | 41.4 [39.4–43.3] |
| Duration Diff (s) | +4.7s | +4.8s | +2.1s | +6.0s |
| $\Delta$Length Bias | 12.9[***] | 14.2[***] | 4.1[*] | **17.3**[***] |

Table 3: Bias Analysis with Statistical Significance Judged by Human and Different LLM Evaluators. Values in brackets are 95% confidence intervals. $\Delta$Bias = difference between top/bottom or long/short preference rates. The detail of computational formula is shown in Appendix C.1.3. [*] $p < 0.05$, [**] $p < 0.01$, [***] $p < 0.001$ (from Permutation Test).

## 6.3 Results and Analysis

### 6.3.1 Overall Performance Landscape

Table 2 presents a stratified yet densely populated leaderboard: while top systems cluster closely together, none surpass the 80-point threshold, indicating substantial room for improvement. Overall, models outperform the average human baseline on *semantic information*, but performance declines for *paralinguistic information* and drops further for *ambient sounds*, highlighting strengths in structured language reasoning alongside persistent limitations in richer auditory contexts. A rubric-level breakdown, provided in Appendix E.1, shows that no single LLM achieves universal leadership; *Security Assessment* exhibits the widest performance variance, whereas most other capabilities display narrower gaps. Consistently, win-rate comparisons, as illustrated in Figure 4, reveal numerous statistically tied outcomes rather than a single dominant system. Together, these findings depict an increasingly competitive landscape, with further progress contingent on advances in multimodal representation, contextual robustness, and domain-specific safety reasoning.

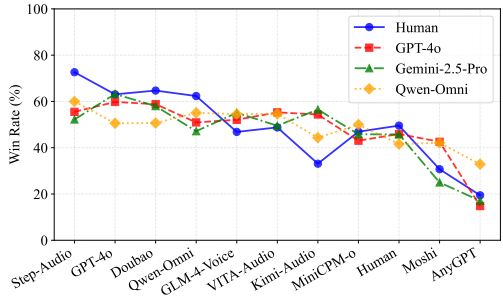

Figure 4: Win rates of different models across different evaluators.[1]

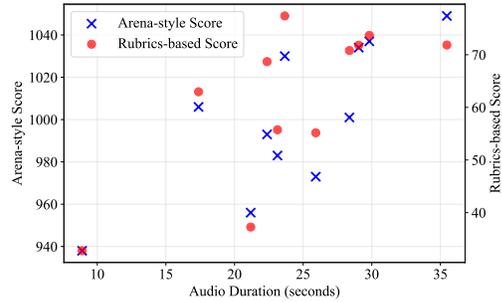

Figure 5: Correlation of audio duration with final Elo scores and Rubric-based evaluation scores.

> **Takeaway 1:** *Top models are strong in overall semantics but limited in specific capabilities like safety reasoning and auditory cues (paralinguistic information and ambient sound processing).*

### 6.3.2 Turn-Level Interaction Patterns

Building on the overall performance picture, we next investigate how models behave across multiple turns. This analysis focuses on two complementary aspects: score dynamics over time and the relationship between output length and answer quality.

**Early Bottleneck vs. Efficiency Drift** Building on the performance landscape above, Table 4 quantifies multi-turn capabilities by tracking **Rubric Score** and **Content Density** across turns. The Rubric-based evaluation measures how well models retain and integrate conversational context at each turn. Two consistent trends emerge. First, **Rubric Score** follows a non-linear trajectory: most systems dip from Turn 1→Turn 2 before recovering at Turn 3. These indicate that the main challenge is not gradual memory decay, but an *early-stage context-accumulation bottleneck*, which means that models struggle to incorporate prior context effectively after the initial turn. Second, **Content Density** declines roughly linearly: later turns contain more tokens yet proportionally less novel information. Taken together, these patterns suggest that models often overcome the early bottleneck by *spending tokens*: regaining coherence in later turns at the expense of informational efficiency.

> **Takeaway 2:** *Models recover from an early-turn quality dip by producing longer responses, trading efficiency for coherence.*

**Length vs. Quality: Minimum Sufficiency Over Verbosity** Figure 5 clarifies that output length is a poor proxy for quality. Extremely short answers underperform, indicating a *minimum viable length* is needed to convey reasoning and evidence; beyond this threshold, additional length may yield diminishing returns. Longer outputs often add redundancy or drift off-topic, without substantive gains. Both human and LLM evaluators should therefore distinguish *informativeness* from *verbosity* and avoid length-based heuristics when assessing model capability.

---

[1]Abbreviations used in Figure 4: GPT-4o (GPT-4o Realtime), VITA-Audio (VITA-Audio-Plus-Vanilla), MiniCPM-o (MiniCPM-o 2.6), Step-Audio (Step-Audio-Chat), Qwen-Omni (Qwen-Omni-Turbo).

| S2S Models | Response Quality | | | Content Density | | |
|---|---|---|---|---|---|---|
| | *T1* | *T2* | *T3* | *T1* | *T2* | *T3* |
| GPT-4o Realtime | 88.5 | 88.3 | 84.9 | 98.8 | 92.4 | 77.5 |
| Doubao | 86.2 | 81.7 | 84.9 | 91.6 | 86.5 | 82.1 |
| GLM-4-Voice | 82.3 | 77.7 | 88.9 | 93.7 | 82.7 | 80.0 |
| VITA-Audio-Plus-Vanilla | 82.5 | 77.9 | 82.8 | 91.8 | 89.8 | 81.0 |
| MiniCPM-o 2.6 | 79.8 | 71.5 | 78.8 | 88.2 | 80.7 | 72.0 |
| Step-Audio-Chat | 81.8 | 79.4 | 77.8 | 86.8 | 77.6 | 80.6 |
| Kimi-Audio | 79.2 | 74.5 | 75.0 | 82.6 | 76.7 | 74.2 |
| Qwen-Omni-Turbo | 82.1 | 74.3 | 75.8 | 92.2 | 69.9 | 69.0 |
| Human | 67.4 | 68.2 | 66.7 | 81.3 | 75.3 | 69.6 |
| Moshi | 54.5 | 46.5 | 45.5 | 70.4 | 65.8 | 64.1 |
| AnyGPT | 67.6 | 55.4 | 64.7 | 58.4 | 30.9 | 29.3 |

Table 4: Turn-level trends in response quality (↑) and content density (↑). T1–T3 denote 1–3 dialogue turns. Darker shades indicate degraded performance. See Appendix D.2 for calculation methods.

> **Takeaway 3:** *After a minimal length for clarity, more tokens often add fluff, not value.*

### 6.3.3 Architectural and Strategic Factors

While the preceding analysis focused on interaction trends, their underlying causes may lie in architectural and training decisions. We next contrast task-specific designs with raw scaling to evaluate their capacity to address early-turn context accumulation and modality coordination.

**Architectural Effects: Task-Specific Designs Outperform Scale**  The turn-level effects align with architectural choices observed in Table 2. Step-Audio-Chat (130B) performs strongly, plausibly due to a specialized design that *transcribes historical turns into text*, conserving the audio context window while leveraging stronger text comprehension. In contrast, many open and commercial baselines encode the entire dialogue history as raw audio within a general multimodal stack, which is not tailored to speech-to-speech demands. Among smaller models, performance variation shows no reliable correlation with parameter count, indicating that *architecture, training strategy, and modality-specific optimizations* matter more than sheer scale. These observations suggest that relieving early context pressure and exploiting modality strengths (e.g., text for long-range history, audio for fresh cues) are more impactful than additional parameters alone.

> **Takeaway 4:** *Paired with large capacity, task-specific designs yield more than scale alone.*

### 6.3.4 Implications: Where Next to Invest

Taken together, the results point to four priorities: (i) **richer multimodal representation & safety robustness**, focusing on capturing paralinguistic and ambient audio information with greater fidelity, while addressing current gaps in semantic dimensions such as *Security Assessment*; (ii) **context management for early-stage bottlenecks**, aimed at mitigating initial-turn context accumulation issues through selective transcription/summarization and task-aware caching; (iii) **task-specific architecture over brute scale**, emphasizing modality-aware designs (e.g., transcribing historical turns to text) that outperform raw-audio pipelines and parameter scaling alone; and (iv) **efficiency-aware output generation**, maintaining *Content Density* beyond a minimal sufficiency threshold while avoiding verbosity.

# 7 Meta-Analysis on Evaluation

## 7.1 Evaluators

### 7.1.1 Human Evaluators vs. LLM-as-a-Judge

Figure 6 compares agreement between LLM-based evaluators and human annotators in Arena-style assessments. In this blind, head-to-head setting, LLM judgments align closely with human preferences when performance disparities are large, but alignment drops sharply when gaps are small. This indicates limited sensitivity to subtle quality differences, which is essential for high-resolution ranking. By contrast, Figure 7 shows that Rubric-based pointwise evaluation yields consistently higher human–LLM agreement across all systems, likely because binary, criterion-specific checks reduce ambiguity. Across both paradigms, Gemini-2.5-pro achieves the highest alignment. These results suggest that LLMs are effective for scalable evaluation when performance gaps are clear or criteria are explicit, but their reliability declines in fine-grained, relative comparisons, underscoring the need for human oversight in high-resolution assessments.

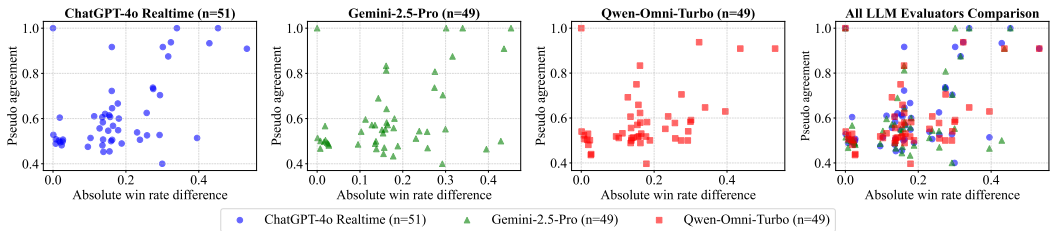

Figure 6: Arena agreements with human evaluators of LLM evaluators.

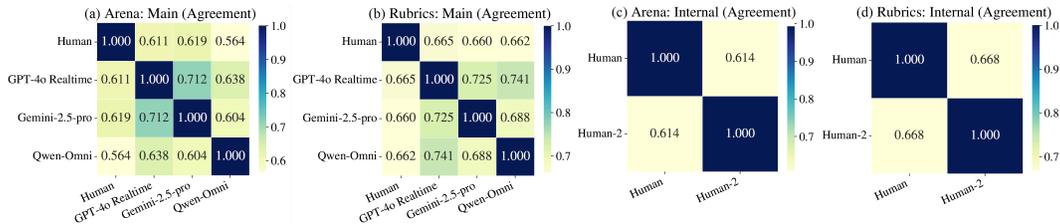

Figure 7: Inter-rater agreement on Arena-style and Rubric-based tasks, including both human-AI (a-b) and human-human (c-d) comparisons.

> **Takeaway 1:** *LLM-as-a-judge performs well with clear gaps or explicit criteria, but human review is essential for precisely assessing fine distinctions.*

### 7.1.2 Bias of LLM-as-a-Judge

Table 3 highlights measurable biases when using LLMs as evaluators. While human judges show minimal positional preference $\Delta$Pos. Bias $= -1.76$, some LLMs exhibit statistically significant biases. For instance, Gemini-2.5-pro favors top-positioned responses (+4.74%), suggesting a susceptibility to positional framing. More notably, all LLMs display a strong and significant length bias, consistently preferring longer responses over shorter ones, with $\Delta$Len. Bias $= +17.3\%$ (*). This bias is substantially greater than that observed in human evaluations (+12.9%, ***), indicating a systematic overvaluation of verbosity by LLM judges. These findings underscore the importance of accounting for structural biases, such as response position and length, when relying on LLMs for automatic evaluation, as such preferences can distort fairness and reliability in comparative assessments.

> **Takeaway 2:** *LLM-as-a-judge shows more obvious biases toward response position and length.*

### 7.2 Evaluation Methods

#### 7.2.1 Arena-style vs. Rubric-based

As shown in Table 2 and Figure 8, our Arena-style and Rubric-based evaluations yield broadly consistent model rankings. This consistency is further substantiated by the internal logical alignment, which demonstrates a high agreement between the two evaluation protocols. While the Arena captures relative user preference, the Rubrics provide a complementary, absolute measure of quality.

Our rubric system's robustness is validated by a bootstrap analysis, as illustrated in Figure 9, where we progressively remove random rubrics, calculate new model rankings from the remaining items, and measure the Spearman correlation ($\rho$) of the new rankings against the original ones. The analysis confirms our rubrics are both **highly self-consistent**, maintaining a strong correlation ($\rho > 0.95$) with the full rubric set (bottom chart), and **externally valid**, showing a stable, high correlation with the Arena ranking (top chart). This strong alignment demonstrates that our analytical rubrics reliably capture the same holistic qualities as pairwise comparisons, justifying our dual-method approach. This consistency is further detailed in Appendix F.1.

> **Takeaway 3:** *Arena-style and Rubric-based evaluations produce consistent rankings, enabling reliable relative and absolute assessments.*

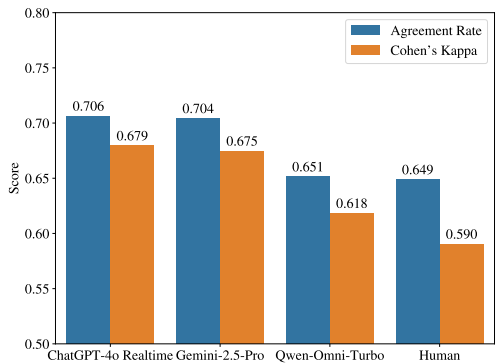

Figure 8: Internal consistency of each evaluator across Arena-style and Rubric-based formats, confirming the reliability of judgments across evaluation methods.

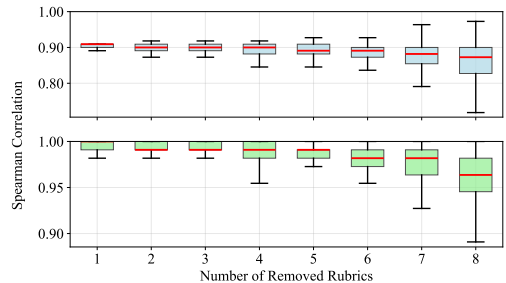

Figure 9: Bootstrap analysis of rubric exclusion (5,000 resamples per condition). The upper panel presents Spearman correlations with Arena-style rankings; the lower panel with original Rubric-based rankings.

#### 7.2.2 Impact of Input Modality on LLM-as-a-Judge Reliability

A critical finding of our study is that the reliability of an LLM-as-a-Judge is highly dependent on the input modality. When evaluating **raw audio**, the LLM's agreement with human evaluators on non-verbal cues (e.g., ambient sounds) is extremely low, with Spearman correlations ($\rho$) dropping to near-zero. However, when these same cues are converted into explicit **annotated transcripts** (e.g., `[laughs]`), the correlation becomes exceptionally high (often $\rho > 0.85$). This indicates that while LLMs struggle to interpret nuanced information from waveforms, they can reliably assess these dimensions once they are textualized, making an annotation-based approach superior for robust automated evaluation. For a detailed breakdown, see Appendix E.3.

> **Takeaway 4:** *LLM-as-a-judge performs poorly on non-verbal audio cues, but becomes reliable when such cues are provided as text annotations.*

#### 7.2.3 Evaluation Pitfalls

Both Arena-style and Rubric-based evaluations face limitations when comparing closely matched models. As shown in Appendix E, small differences in ELO rating often correspond to negligible or unstable win-rate gaps, and in some cases, even inversely correlate. Figure 10 further illustrates that when score differences (Arena-style or Rubric-based) are minimal, increasing the number of comparisons does little to clarify superiority. Both methods yield stable and reliable distinctions only when performance gaps are sufficiently large. These findings caution against over-interpreting small score or ranking differences, especially in high-variance, near-parity settings.

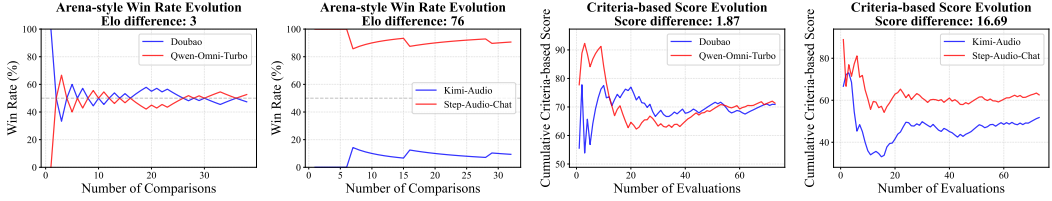

Figure 10: Win rates and rubric scores of model pairs of low or high score difference.

> **Takeaway 5:** *In Arena-style or Rubric-based evaluations, only large gaps give reliable results.*

# 8 Conclusion

We present MTalk-Bench, a comprehensive benchmark designed to evaluate multi-turn S2S LLMs across three critical dimensions: semantic understanding, paralinguistic expression, and ambient acoustic quality. By conducting large-scale evaluations with both human judges and LLM-based assessments, we systematically uncover the current capabilities and limitations of state-of-the-art models. Our analysis reveals notable strengths in short-turn semantic comprehension, but also exposes persistent challenges in maintaining contextual coherence over longer dialogues, generating expressive prosody, and achieving conversational efficiency. These findings underscore a pressing need for next-generation models to move beyond mere content correctness, toward more concise, context-sensitive, and naturally expressive spoken interactions that better reflect human communication.

# A    Framework Details of MTalk-Bench

## A.1    User-voted Scenarios

The architecture of MTalk-Bench is grounded in a data-driven, two-part methodology designed to establish its evaluation scenarios and capability dimensions. This approach systematically combines large-scale user preference surveys with established theoretical frameworks from linguistics and computer science. The primary goal is to ensure the benchmark's ecological validity by focusing on communication contexts and model abilities that users deem most critical, while also maintaining a comprehensive and theoretically sound structure for evaluation.

**Participants.**    Over 80 university students participate in the survey studies described in this section. All participants engage with the tasks voluntarily and are informed of the study's purpose. The detailed instructions provided to the participants for each survey task are elaborated in the corresponding subsections below.

### A.1.1    Scenario Selection

To ensure the relevance of our benchmark, we first identify a set of high-frequency communication scenarios pertinent to S2S LLMs. The process begin with an extensive literature review across communication studies, sociology, and Human-Computer Interaction (HCI) to compile a broad list of real-world contexts Gumperz (1982); Clark (1996); Schegloff (2007).

Subsequently, we conduct a pairwise comparison survey to empirically rank these scenarios. In each trial, participants are presented with two scenarios side-by-side (e.g., "Scenario X vs. Scenario Y") and are instructed to select the one they believed to be more likely to occur in a real-world setting. The instructions explicitly emphasize that choices should be based on perceived frequency or plausibility, not personal preference or desirability. After each selection, a new pair of scenarios is randomly sampled for evaluation. This pairwise comparison method is highly effective for eliciting robust preference data while mitigating biases common in direct rating scales Thurstone (1927); David (1988).

The survey yield an empirical ranking of scenarios based on their Elo scores, quantifying their perceived relevance for future interactions with speech-based AI agents (see Table 6). To construct the benchmark, we directly select the **top nine scenarios** from this ranking. This data-driven approach ensures that MTalk-Bench prioritizes contexts that users identify as most significant. The resulting set also spans a diverse range of communicative functions, from institutional and professional interactions to personal and socio-emotional conversations. The nine scenarios selected for MTalk-Bench are:

1. **Family and Domestic Communication** (e.g., coordinating household tasks, family scheduling, managing smart home devices via voice)

2. **Health and Medical Communication** (e.g., initial symptom checking, virtual health assistant consultations, medication reminders, accessing medical information)

3. **Institutional Inquiry** (e.g., querying government services, basic legal information retrieval, financial account inquiries, university helpdesks)

4. **Educational Communication** (e.g., AI-powered tutoring, language learning applications, interactive educational Q&A, voice-guided tutorials)

5. **Workplace Communication** (e.g., meeting dictation and summarization, collaborative task management via voice, professional information lookup, job interview practice)

6. **Entertainment Communication** (e.g., interacting with voice-controlled games, generating stories or scripts via speech, controlling media playback, interactive audio experiences)

7. **Casual Interaction** (e.g., open-domain social chat, companionship with AI, storytelling, expressing feelings and receiving empathetic responses)

8. **Psychological Communication** (e.g., AI coaches for well-being, guided mindfulness exercises, initial mental health support and resource navigation)

9. **Service-Oriented Communication** (e.g., customer service inquiries, booking appointments, technical support, retail assistance, travel planning)

| Capability | Elo Score |
|---|---|
| Understanding & Memory | 1028 |
| Reasoning & Execution | 1027 |
| Interaction Strategy | 1020 |
| Paralinguistic Generation | 1019 |
| Pragmatics Culture | 1018 |
| Dynamic Reverberation Compensation | 1018 |
| Real-Time Voice Quality Restoration | 1001 |
| Ambient sound perception & adaptation | 998 |
| Multi-party interaction understanding | 998 |
| Security Assessment | 993 |
| Continual Learning & Adaptive Semantic | 989 |
| Modeling Cross-lingual & Conceptual Generalization | 989 |
| Paralinguistic Comprehension | 987 |
| Contextual Adaptation Capability | 986 |
| Semantic Robustness | 986 |
| Turn-taking & Interruption Handling | 984 |
| Dialect & Accent Robustness | 984 |
| Low Latency Response | 966 |

Table 5: Elo rankings of S2S-LLM capabilities from a pairwise preference survey. Higher scores indicate greater perceived importance for near-future AI agents.

| Scenario | Elo Score |
|---|---|
| **Family and Domestic Communication** | **1091** |
| **Health and Medical Communication** | **1070** |
| **Institutional Inquiry** | **1048** |
| **Educational Communication** | **1045** |
| **Workplace Communication** | **1029** |
| **Entertainment Communication** | **1027** |
| **Casual Interaction** | **1027** |
| **Psychological Communication** | **1017** |
| **Service-Oriented Communication** | **1008** |
| Public Discourse and Interaction | 1007 |
| Diplomatic and International Relations Communication | 1003 |
| Marketing and Customer Relationship Management Communication | 1001 |
| Negotiation and Conflict Resolution | 995 |
| Sports and Competitive Communication | 980 |
| Religious and Spiritual Communication | 979 |
| Intercultural and Linguistic Communication | 974 |
| Public Affairs and Emergency Response Communication | 960 |
| Military and Tactical Discussion | 950 |
| Tourism and Local Interaction Communication | 906 |

Table 6: Elo rankings of evaluation scenarios from a pairwise preference survey. Participants selected the scenario more likely to involve a speech-based AI agent. The top nine (bolded) are selected for MTalk-Bench.

### A.2 User-voted Capability

The capability structure are defined through a similar synthesis of literature-based frameworks and empirical validation. We begin by establishing a three-dimension evaluation structure, a common practice in designing comprehensive benchmarks that allows for a multi-faceted assessment of model abilities Wang et al. (2020); Yu et al. (2023). In the context of multi-turn spoken dialogue, a successful interaction depends not only on the agent's ability to continuously comprehend and reason about semantic content, but also critically on its capacity to interpret and generate the paralinguistic cues (e.g., emotion, intent) that shape the conversational dynamic. Furthermore, robustness in real-world settings necessitates adapting to complex acoustic conditions and navigating multi-speaker challenges. A comprehensive assessment of these three dimensions is therefore essential for holistically measuring the proficiency of a spoken dialogue system. This framework, informed by tiered models of communication Levelt (1989), divides capabilities into:

- **Type I: Semantic Information**, focusing on the understanding and generation of literal content.
- **Type II: Paralinguistic Information**, assessing the handling of non-lexical vocal cues like emotion and prosody Scherer (1986).
- **Type III: Ambient Sound**, evaluating adaptation to the acoustic context, such as background noise or multiple speakers Bregman (1990).

With this structure in place, we compile an extensive list of candidate capabilities within each dimension, drawing from foundational work in linguistics Grice (1975), HCI, and recent speech processing research. We then conduct a single, unified pairwise comparison survey. Participants are instructed to imagine interacting with a voice-based intelligent agent and evaluate the relative importance of its potential capabilities. In each trial, two capability dimensions are presented side-by-side (e.g., "Dimension 1 vs. Dimension 10"), and participants select the one they consider more critical or useful for such an agent to possess, based on their personal judgment. This process repeats with randomly generated pairs until all comparisons are completed. Such preference-based surveys are a valid and reliable method for evaluating complex system capabilities Saaty (2008).

The final set of capabilities for MTalk-Bench is determined by selecting the **top-ranked items from within each of the three predefined dimensions**, ensuring that the benchmark maintains a balanced focus across semantic, paralinguistic, and ambient dimensions. The resulting Elo rankings are presented in Table 5, and the detailed breakdown of the selected capabilities is provided in the subsequent paragraphs.

**Semantic Information**   This dimension assesses the model's ability to comprehend, reason about, and strategically manage the explicit semantic content of spoken dialogue. The evaluation is structured into the following capabilities:

- **Understanding & Memory:** Assesses the model's ability to retain and accurately utilize information from the dialogue history.
  - *Context Consistency:* Measures the ability to maintain logical and factual coherence across multiple conversational turns.
  - *Semantic Disambiguation:* Evaluates the capacity to resolve ambiguity in words or phrases by leveraging contextual clues.
  - *Content Reforming:* Tests tasks such as summarization, rephrasing, and key information extraction based on prior conversation.
- **Reasoning & Execution:** Measures the model's proficiency in complex cognitive tasks that require logical inference and planning.
  - *Task Planning:* Assesses the decomposition of high-level requests into a sequence of actionable, logical steps.
  - *Logical & Reasoning:* Probes commonsense, deductive, and inductive reasoning capabilities within a conversational context.
- **Interaction Strategy:** Assesses the model's effectiveness in managing conversational dynamics and flow.
  - *Dialogue Management:* Evaluates the handling of conversational phenomena like interruptions, topic shifts, and proactive engagement.
  - *Error Handling:* Measures the ability to gracefully manage and recover from ambiguous, incomplete, or unanswerable user queries.
- **Security Assessment:** Probes the model's alignment with safety protocols and its ability to act responsibly.
  - *Bias Detection:* Assesses the capability to identify and refuse to perpetuate harmful stereotypes present in user speech.
  - *Safety Risk Detection:* Measures the ability to recognize and appropriately respond to content that is inappropriate or indicates potential harm.
- **Pragmatics & Culture:** Tests the model's grasp of nuanced communication that extends beyond literal meaning.
  - *Nonliteral Understanding:* Evaluates the comprehension of sarcasm, humor, metaphors, and other forms of figurative language.
  - *Cultural Fit:* Assesses the model's capacity to adapt its responses to diverse cultural norms, etiquette, and contextual expectations.

**Paralinguistic Information**   This dimension evaluates the model's ability to interpret and generate non-lexical vocal cues that convey emotion, intent, and identity. Capabilities are divided into comprehension and generation tasks.

- **Paralinguistic Comprehension:** Focuses on the model's ability to interpret the non-lexical information embedded in speech signals Trager (1958).
  - *Emotion Detection:* Identifying the speaker's affective state (e.g., joy, anger, sadness) from vocal prosody.
  - *Paralinguistic Signal Recognition:* Interpreting cues such as stress, intonation, speech rate, and pitch to understand emphasis and intent.
  - *Speaker Recognition:* Differentiating between or identifying speakers based on their unique vocal characteristics.
- **Paralinguistic Generation:** Assesses the model's proficiency in producing speech with specific, controlled expressive qualities.

- *Emotional Speech:* Synthesizing speech that convincingly conveys a target emotion.
- *Paralinguistic Signal Generation:* Producing speech with deliberate control over prosody, rhythm, and intonation.
- *Expressive Modeling:* Emulating a specific speaker's style, accent, or vocal mannerisms.

**Ambient Sound**  This dimension measures the model's robustness and contextual awareness in realistic, non-sterile acoustic environments Rabiner and Juang (1993).

- **Ambient Sound Perception:** Tests the model's ability to process and reason about its acoustic surroundings.
  - *Ambient Sound Understanding:* Identifying and correctly interpreting non-speech sounds (e.g., a ringing phone, a passing siren, music).
  - *Noise Robustness:* Maintaining high performance in speech recognition and semantic understanding despite the presence of background noise.
  - *Ambient Cue Reasoning:* Leveraging identified background sounds to make logical inferences about the user's environment or situation.
- **Multi-party Interaction:** Evaluates performance in conversations involving multiple participants.
  - *Speaker-aware Modeling:* Attributing speech segments to the correct speaker in a multi-talker scenario (i.e., speaker diarization).
  - *Interaction Coherence:* Managing turn-taking and maintaining a coherent conversational flow between multiple, potentially overlapping speakers.

### A.3 Scenario-Capability Mapping

To ensure the constructed data is reasonable, we map each scenario's communicative demands to specific capabilities in our taxonomy. This involves analyzing real-world requirements along three defined dimensions.

**Family and Domestic Communication**  This scenario encompasses common household interactions, such as coordinating daily routines, managing family schedules, or engaging with smart home devices. These conversations often involve overlapping speech, familiar speakers, and emotionally charged content.

- **Semantic Information:** *Comprehension & Memory* is the most critical capability in this domain, as effective communication depends on tracking prior context, shared responsibilities, and time-sensitive instructions.
- **Paralinguistic Information:** *Paralinguistic Comprehension* is essential for interpreting subtle emotional cues and speaker intent, which are particularly prominent in family dynamics.
- **Ambient Sound:** Given the acoustic complexity of home environments—ranging from kitchen noise to television or children playing—*Ambient Sound Understanding* plays a key role in maintaining intelligibility and contextual awareness.

**Health and Medical Communication**  This scenario involves sensitive and high-stakes interactions like symptom checking, virtual consultations, and medication management.

- **Semantic Information:** Effective health communication requires conveying complex medical information clearly and accurately, which directly impacts patient outcomes (Street Jr et al., 2009). This aligns with our *Reasoning & Execution* capability, the most critical semantic skill in this scenario.
- **Paralinguistic Information:** Building patient trust depends on recognizing emotional cues in speech (e.g., anxiety or pain) (Roter et al., 1988). Given its diagnostic value, *Paralinguistic Comprehension* is the most emphasized capability here.
- **Ambient Sound:** Medical environments are acoustically complex, with both background noise and clinically relevant sounds like alarms (Stowell et al., 2015). To function reliably, a model must distinguish speech from ambient sounds and extract key cues—captured by our *Ambient Sound Understanding* capability.

**Institutional Inquiry**   This scenario captures formal, information-driven exchanges with institutional bodies such as government agencies, banks, or university helpdesks. These interactions often involve rigid procedural structures, high stakes, and variable acoustic environments.

- **Semantic Information:** *Interaction Strategy & Intelligence* is the most critical capability here, as institutional dialogues require managing complex question-answer flows, clarifying ambiguous requests, and adhering to procedural turn-taking (Clark, 1996; Schegloff, 2007).

- **Paralinguistic Information:** *Paralinguistic Comprehension* enables the system to detect caller frustration, hesitation, or urgency—essential for regulating tone and adapting response strategy in service-oriented settings (Scherer, 1986).

- **Ambient Sound:** With the highest score in *Ambient Sound Understanding* across all scenarios, this setting reflects environments such as call centers, where background chatter, typing, or announcements can degrade communication (Stowell et al., 2015). Robust handling of acoustic interference is essential for intelligibility and task success.

**Educational Communication**   This scenario involves learning-oriented settings such as one-on-one tutoring, language acquisition, and instructional dialogues. Effective communication in this domain requires both conceptual clarity and pedagogical sensitivity.

- **Semantic Information:** *Reasoning & Execution* is the most critical capability, as educational interactions rely on the model's ability to explain concepts, correct misunderstandings, and adapt explanations to a learner's developmental stage (Gan et al., 2023).

- **Paralinguistic Information:** A strong emphasis on *Paralinguistic Generation* enables the model to deliver responses with appropriate intonation, encouragement, and clarity—factors known to enhance learner engagement and comprehension (Nass and Brave, 2005).

- **Ambient Sound:** Collaborative learning often involves multiple speakers—teachers, students, or peers—sometimes in noisy environments. *Multi-party Interaction* is thus essential for tracking speaker turns and maintaining conversational coherence (Lave and Wenger, 1991).

**Workplace Communication**   This scenario simulates professional settings such as summarizing meetings, managing projects, and engaging in structured work-related dialogue. These interactions are typically information-dense, time-sensitive, and often involve multiple stakeholders.

- **Semantic Information:** *Comprehension & Memory* is the most critical capability in workplace contexts, as models must retain and organize key information across turns to support activities such as minute-taking, scheduling, and decision tracking (Hu et al., 2025).

- **Paralinguistic Information:** *Paralinguistic Comprehension* plays a vital role in interpreting professional tone, urgency, or disagreement—essential elements in navigating meetings and negotiations (Scherer, 2003).

- **Ambient Sound:** *Multi-party Interaction* is the dominant ambient-related capability, as workplace communication frequently involves multiple participants, overlapping speech, and dynamic turn-taking patterns (Janin et al., 2003).

**Entertainment Communication**   This scenario encompasses leisure-oriented use cases such as interactive storytelling, voice-controlled games, and media navigation. These tasks demand creativity, contextual awareness, and strong expressive abilities.

- **Semantic Information:** *Pragmatics & Culture* is the most important capability, as entertainment-focused interactions often rely on understanding humor, cultural references, and nonliteral language (Attardo, 1994).

- **Paralinguistic Information:** *Paralinguistic Generation* plays a central role in rendering character voices, maintaining narrative engagement, and producing expressive delivery in storytelling and games (Skerry-Ryan et al., 2018).

- **Ambient Sound:** *Ambient Sound Understanding* is critical for responding to or integrating with background music, sound effects, or ambient media cues—hallmarks of immersive entertainment experiences (Bregman, 1990).

**Casual Interaction** This scenario includes informal, open-ended conversations intended to foster companionship, maintain social presence, or simulate human-like small talk. These dialogues often involve fluid topic shifts and social nuance.

- **Semantic Information:** *Reasoning & Execution* is the most critical semantic capability, as maintaining engaging, natural conversations depends on commonsense reasoning and flexible topic handling (Sap et al., 2019).

- **Paralinguistic Information:** *Paralinguistic Comprehension* plays a key role in interpreting tone, sarcasm, enthusiasm, or hesitation—features that help the model mirror conversational affect and maintain rapport (Scherer, 1986).

- **Ambient Sound:** With casual settings often occurring in public or group environments, *Multi-party Interaction* is essential for managing turn-taking, speaker identification, and overlapping dialogue (Janin et al., 2003).

**Psychological Communication** This scenario involves emotionally sensitive interactions, including mental health coaching, mindfulness guidance, and low-stakes crisis support.

- **Semantic Information:** *Pragmatics & Culture* is central to this domain, enabling models to deliver responses that are empathetic, nonjudgmental, and culturally attuned (Rogers, 1951).

- **Paralinguistic Information:** *Paralinguistic Comprehension* supports the accurate perception of emotional cues, such as stress or vulnerability, which is essential for building rapport and responding appropriately (Scherer, 2003).

- **Ambient Sound:** *Multi-party Interaction* plays a supporting role in sessions involving caregivers or supportive peers, requiring the model to manage overlapping input with sensitivity.

**Service-Oriented Communication** This scenario involves task-driven interactions such as customer support, appointment scheduling, or travel booking.

- **Semantic Information:** *Security Assessment* is the most critical capability, as interactions often involve sensitive personal or financial data that must be handled safely and responsibly (Cavoukian, 2009).

- **Paralinguistic Information:** *Paralinguistic Generation* ensures that responses are delivered clearly and professionally, supporting consistent tone and user trust (Nass and Brave, 2005).

- **Ambient Sound:** *Ambient Sound Understanding* is essential for maintaining robustness in acoustically challenging environments like call centers or public venues (Stowell et al., 2015).

Specific data information for all scenarios and capability dimension mapping can be found in Table 7.

### A.3.1 Methodological Soundness and Rationale

The design of MTalk-Bench, encompassing both the selection of communication scenarios and the definition of evaluative capability dimensions, is underpinned by a commitment to methodological soundness, user-centered principles, and data-driven decision-making. This approach ensures the benchmark's relevance, comprehensiveness, and robustness for evaluating S2S LLMs.

The methodology for developing MTalk-Bench is deliberately chosen to ensure its relevance and rigor. By grounding both scenario and capability selection in empirical user preferences obtained via pairwise comparison surveys Thurstone (1927); Saaty (2008), we ensure the benchmark possesses strong ecological validity. This data-driven approach prevents an over-reliance on researcher intuition and aligns the evaluation criteria with real-world user expectations.

## B  Benchmark Construction Details

This appendix provides a detailed technical explanation of the methodologies for generating the MTalk-Bench dialogue instances.

### B.1 Constructing Raw Textual Dialogue

The construction of the Focus-Semantics Dialogue Dataset involves a synergistic approach combining scripted LLM-based generation with subsequent human refinement to ensure data quality, relevance, and balanced coverage. The process utilizes three distinct scripts:

1. **Script 1 (Dialogue Generation):** Employ a large-scale, instruction-tuned LLM to generate initial multi-turn dialogues based on a specified scenario and primary capability. A key constraint is ensuring the final turn's response depends on context from earlier turns to test multi-turn reasoning.

2. **Script 2 (Dimension Labeling):** Utilize a second LLM to annotate each generated dialogue with all potentially relevant capabilities from our taxonomy.

3. **Script 3 (Primary Dimension Inference):** Leverage a computationally efficient model to infer the single most prominent evaluation dimension from the labels generated by Script 2.

This automated generation is followed by a rigorous human-in-the-loop refinement process. Initially, over 1500 candidate dialogues are generated. A team of trained annotators reviews these candidates, filtering them for logical coherence, dialogue naturalness, and the unambiguous testability of the intended capability. This multi-stage quality control process results in the retention of approximately 19% of the initial set. Any discrepancies between the intended primary dimension (input to Script 1) and the inferred one (output from Script 3) are manually resolved. This process guarantees a balanced distribution of at least 10 high-quality instances per targeted capability. This iterative refinement is crucial for creating datasets that are both scalable and reliable for benchmarking advanced dialogue systems Gao et al. (2023).

### B.2 Tag Design for Paralinguistic and Ambient Sound

#### B.2.1 Focus-Paralinguistic Dataset Construction

The design of paralinguistic tags in MTalk-Bench is grounded in established principles from affective computing and speech science to ensure that our evaluation is both meaningful and robust. The primary goals are to assess a model's ability to perceive, interpret, and generate non-lexical vocal cues that are critical for human communication Picard (1997). Our tags are designed based on principles of **communicative relevance** and **perceptual distinctiveness**, drawing from large-scale analyses of vocal expression Schuller et al. (2013).

The tags and their distribution across scenarios are structured to test two primary functions: the comprehension of user input and the controlled generation of model output.

**Paralinguistic Comprehension (53 instances).** This function evaluates the model's ability to interpret vocal cues in the user's speech.

- **Emotion Detection (22 instances):** Assesses the ability to identify affective states from voice. This is tested using descriptive tags on user input (e.g., `<anxious tone>`). For instance, the *Psychological Communication* scenario features a high concentration of these cues (4 instances) to directly evaluate a model's empathetic understanding.

- **Paralinguistic Signal Recognition (23 instances):** Focuses on interpreting prosodic features that govern rhythm, stress, and intonation Cutler and Ladd (1997). We use tags like `<slow pace>` or `<hesitant tone>`. These are particularly dense in the *Health* scenario (6 instances), reflecting the clinical importance of interpreting subtle vocal cues like vocal strain.

- **Speaker Recognition (8 instances):** Tests the ability to recognize speaker identity or infer characteristics from vocal traits Kinnunen and Li (2010). This is evaluated through dialogue context involving multiple speakers or explicit style descriptions.

**Paralinguistic Generation (37 instances).** This function evaluates the model's ability to control its own vocal expression according to specific instructions.

- **Emotional Speech (11 instances):** Assesses the ability to synthesize speech with a specified emotion. This is directed via instructional tags (e.g., `<respond in a reassuring`

`tone>`). For example, in the *Psychological Communication* scenario, the model is explicitly tested on its ability to generate empathetic speech.

- **Paralinguistic Signal Generation (15 instances):** Focuses on producing speech with specified prosodic characteristics. Instructions like `<speak slowly and clearly>` are used to test fine-grained control over the model's vocal output, a key feature in scenarios like *Family and Domestic Communication* (3 instances).

- **Expressive Modeling (11 instances):** Tests the ability to adopt a specific vocal persona or style. This is heavily tested in scenarios like *Institutional Inquiry* (5 instances) and *Service-Oriented Communication* (4 instances), simulating tasks where adopting a consistent brand persona is required.

This scenario-aware distribution of paralinguistic tags allows for a fine-grained evaluation of how well models adapt their understanding and expression to different social and functional contexts.

### B.2.2 Focus-Ambient Sound Dataset Construction

This dimension evaluates a model's ability to maintain robust and context-aware communication in realistic acoustic environments. The design of our ambient sound tags is guided by the principle of **ecological validity**—representing a wide range of common real-world acoustic events that can impact a conversation Gemmeke et al. (2017). These tags are structured to test two core capabilities:

- **Ambient Sound Perception & Adaptation**: This capability is tested using tags that describe the acoustic scene. These are further divided into:

  - *Discrete Events*: Short, distinct sounds that may require a direct response or inference (e.g., `<phone ringing>`, `<door slams>`, `<dog barking>`).
  - *Continuous Noise*: Background sounds that test a model's signal processing robustness and ability to adapt its output, such as speaking more loudly (e.g., `<cafe chatter>`, `<street traffic>`).
  - *Signal Integrity Issues*: Events that directly corrupt the speech signal, testing a model's ability to handle missing information (e.g., `<speech obscured by cough>`, `<static interruption>`).

  The *Institutional Inquiry* scenario is composed entirely of these challenges (10 instances), simulating interactions in noisy public spaces.

- **Multi-Party Interaction Tracking**: Beyond simple background noise, we test a model's ability to understand complex social dynamics. This is crucial for real-world deployment where multiple speakers are common. We use tags to describe the conversational flow and turn-taking events, informed by principles of conversation analysis Stolcke et al. (2000). Examples include `<Speaker B interjects>`, `<User turns to address Speaker C>`, and `<Two people talking in background>`. Scenarios with high social complexity, such as *Casual Interaction* (6 instances) and *Workplace Communication* (5 instances), feature a high density of these multi-party tags to evaluate a model's ability to track who is speaking and what the social implications are.

By systematically incorporating these diverse acoustic and interactional challenges, MTalk-Bench provides a comprehensive testbed for evaluating the environmental robustness and social intelligence of S2S models.

### B.3 Audio Generation Pipeline

This section provides further details on the audio synthesis pipeline.

**Human Recording Protocol**  To capture naturalistic human speech, we recruit native English speakers from English-speaking countries via the Amazon Mechanical Turk (MTurk) platform. For dialogue instances targeting **Semantic** capabilities, participants are instructed to read the provided text with neutral prosody and clear articulation. In contrast, for the **Paralinguistic** instances, participants receive explicit instructions derived from the dialogue metadata tags (e.g., `<gentle tone>`, `<angry>`, `<whispering>`). These instructions guide them to produce specific vocal affects and prosodic variations, ensuring the resulting audio faithfully represents the intended paralinguistic intent, which is a critical channel of human communication Schuller et al. (2013).

| Scenario | Semantic Information | | | | | Paralinguistic Information | | Ambient Sound | |
|---|---|---|---|---|---|---|---|---|---|
| | Comprehension & Memory | Reasoning and Task Execution | Security Assessment | Pragmatic and Cultural Competence | Interaction Strategy and Intelligence | Paralinguistic Comprehension | Paralinguistic Generation | Ambient Sound Understanding | Multi-party Interactive Understanding |
| Family and Domestic Communication | 3 | 2 | 2 | 1 | 1 | 5 | 4 | 6 | 3 |
| Health and Medical Communication | 2 | 4 | 1 | 1 | 2 | 8 | 2 | 6 | 4 |
| Institutional Inquiry | 1 | 3 | 1 | 1 | 4 | 4 | 6 | 10 | 0 |
| Educational Communication | 0 | 3 | 3 | 1 | 3 | 5 | 6 | 5 | 5 |
| Workplace Communication | 4 | 0 | 2 | 1 | 3 | 7 | 3 | 5 | 5 |
| Entertainment Communication | 3 | 1 | 1 | 3 | 2 | 5 | 5 | 7 | 3 |
| Casual Interaction | 0 | 6 | 2 | 2 | 2 | 10 | 2 | 6 | 6 |
| Psychological Communication | 0 | 1 | 2 | 4 | 3 | 6 | 4 | 6 | 4 |
| Service-Oriented Communication | 1 | 2 | 3 | 2 | 1 | 3 | 5 | 7 | 2 |
| Total | 14 | 22 | 17 | 16 | 21 | 53 | 37 | 58 | 32 |

Table 7: Combined scenario-capability mapping across semantic, paralinguistic, and ambient benchmarks

To match the diversity of real-world communication contexts, we organize recruitment and recording in **nine separate MTurk batches**, each corresponding to a specific scenario from our dataset. For each batch, the **Semantic** dialogue instances and their corresponding **Paralinguistic** and **Ambient** variants are grouped into a single task. Each task is priced at **USD 0.40** per completed set.

The MTurk task is advertised under the title: *Read Subtitled Dialogues and Record Audio (Native English Speakers Only)"*, with the following description: *We are looking for native English speakers to read and record dialogues. The audio will be used for linguistic research."*

Workers were required to meet the following qualifications:

1. Masters Qualification on Amazon Mechanical Turk.

2. HIT Approval Rate (%) greater than 65 across all requesters' HITs.

3. Location must be one of AUSTRALIA (AU), CANADA (CA), IRELAND (IE), NEW ZEALAND (NZ), UNITED KINGDOM (GB), or UNITED STATES (US).

All submitted recordings are **manually reviewed** by our research team. Audio that fails to meet our quality standards—such as lack of required paralinguistic expression, unclear articulation, excessive background noise, or overly low volume—is rejected, and the corresponding task is reposted. We conduct three iterative rejection-and-recollection rounds:

- **Round 1:** 25 rejections (3 due to incomplete/mistaken reading, 22 due to missing paralinguistic expression).

- **Round 2:** 11 rejections (1 due to incomplete/mistaken reading, 10 due to missing paralinguistic expression).

- **Round 3:** 4 rejections (all due to missing paralinguistic expression).

After the third round, all audio passes quality control. The final dataset consists entirely of recordings that satisfy our semantic, paralinguistic, and acoustic criteria.

**Voice Conversion Model** Generating audio for specialized voice profiles, such as those of children or the elderly, is often impractical via direct data collection. To overcome this challenge, we employ Seed-VC Liu (2024), a state-of-the-art voice conversion framework that leverages self-supervised speech representations. A key advantage of Seed-VC is its text-free and reference-free nature, allowing for the transformation of vocal timbre without requiring corresponding text transcriptions or parallel reference recordings from the target speaker. This methodology allows us to synthesize realistic child and elderly voices while preserving the original prosody and emotional content captured during the human recording phase.

**Ambient Sound Curation and Integration** To simulate the diverse and often noisy acoustic environments of real-world interactions, we integrate background ambient sounds into the dialogue recordings. A comprehensive library of background audio clips is curated from established open-source repositories, including Freesound Font et al. (2013), Pixabay Sound Effects, and public datasets on GitHub such as FSD50K Fonseca et al. (2022). Selected clips, representing environments like bustling cafés, urban traffic, and quiet offices, undergo post-processing for volume normalization.

The mixing process is conducted manually using professional audio editing tools such as Adobe Audition Adobe Inc. (2023) and Jianying Pro Shenzhen Lianmeng Technology Co., Ltd. (2023), ensuring precise alignment and seamless blending between the speech and the background environment. This manual approach allows to achieve a natural and realistic acoustic scene.

Following integration, a final human review is performed to verify that each mixed recording realistically reflects its intended scenario, avoiding unnatural overlaps, masking of critical speech segments, or inconsistencies with the scene description. This quality-control step guarantees that the benchmark can effectively test a model's robustness to background noise and its ability to comprehend speech in ecologically valid settings.

## C   Evaluation Protocol Details

This section details our evaluation framework, including the core principles of the Arena and Rubrics methods, the Elo rating system, and the hierarchical rubric design.

### C.1   Arena-style Evaluation Protocol

#### C.1.1   Arena Interface and Sampling Logic

In the Arena evaluation interface, annotators are randomly assigned to one of the three benchmark dimensions—*Semantic Information*, *Paralinguistic Information*, or *Ambient Sound*. Within the selected dimension, the annotator is presented with audio outputs from model pairs that have the closest current Elo scores, thereby prioritizing comparisons between similarly performing systems. This dynamic pairing strategy ensures high-resolution ranking where it matters most, and reflects practical user preference through fine-grained matchups. A sample interface used for arena evaluation is shown in Figure 13.

Each Arena task includes:

- Description of the assessment capabilities for this round
- A system prompt and user input (in audio form)
- Two model responses (in audio form)
- The button to select which model performs better
- The text box where users explain for their choice

#### C.1.2   Elo Rating System

To obtain a comparative ranking of S2S models in the S2S-Arena framework, we adapt an Elo rating system, originally developed for chess ranking, to aggregate results from pairwise model comparisons. This system provides a robust method for handling the dynamic nature of model comparisons and ensures stable rankings even with varying numbers of evaluations per model pair.

**Initialization and Setup**   Each model is assigned an initial Elo score of 1000, which will be updated based on the outcomes of pairwise comparisons. Let model $A$ and model $B$ be compared on the same evaluation instance, with each pair receiving one of the following outcomes based on human or LLM judgment: $A$ wins over $B$ ($S_A = 1$, $S_B = 0$), or $B$ wins over $A$ ($S_A = 0$, $S_B = 1$).

**Score Update Rule**   Let $R_A$ and $R_B$ denote the current Elo scores of models $A$ and $B$, respectively. The expected win probability for $A$ is computed as:

$$E_A = \frac{1}{1 + 10^{(R_B - R_A)/400}}, \quad E_B = 1 - E_A$$

The Elo scores are then updated using:

$$R_A' = R_A + K(S_A - E_A)$$

$$R'_B = R_B + K(S_B - E_B)$$

where $K$ is a constant controlling the update rate. We use $K = 4$ in our experiments, following common practice in Elo-based evaluation systems.

**Aggregation and Ranking** The final Elo score of each model is computed after all pairwise comparisons are completed across evaluation instances. Models are then ranked in descending order of their final Elo scores, providing a comprehensive ranking that reflects their relative performance across all evaluation dimensions.

### C.1.3 Statistical Inference Methods for Arena Bias Analysis

We define below the metrics and statistical tests used for analyzing position and length biases in S2S model preferences.

**Preference Rate (TPR, BPR, LPR, SPR)** For a given preference condition (e.g., top position), we define the preference rate as:

$$\text{Preference Rate} = \frac{n_{\text{preferred}}}{N}$$

where $n_{\text{preferred}}$ is the number of times the preferred category (e.g., top or long) is selected, and $N$ is the total number of evaluation instances.

**Bias Score (Difference in Preference)** To quantify directional bias, we compute the difference in preference rates between two competing categories:

$$\Delta_{\text{bias}} = p_1 - p_2$$

where $p_1$ and $p_2$ are the preference rates for the two categories, such as top vs. bottom (for position bias) or long vs. short (for length bias). A positive $\Delta_{\text{bias}}$ indicates a bias towards category 1.

**Confidence Interval (Wilson Score)** The 95% confidence interval for a preference rate $p = \frac{x}{n}$ is calculated using the Wilson Score Interval:

$$\hat{p} = \frac{x + \frac{z^2}{2}}{n + z^2}, \quad z = 1.96$$

$$\text{half-width} = \frac{z \cdot \sqrt{\frac{x(n-x)}{n} + \frac{z^2}{4}}}{n + z^2}$$

$$\text{CI}_{95\%} = \hat{p} \pm \text{half-width}$$

This interval is more accurate than the normal approximation, especially when $p$ is near 0 or 1 or when $n$ is small.

**Permutation Test for Significance of Bias** To assess whether the observed bias $\Delta_{\text{obs}}$ is statistically significant, we conduct a non-parametric permutation test:

1. Combine all preference labels (e.g., "top" and "bottom") into a single set of size $N$.

2. Randomly shuffle the labels and reassign them into two groups of sizes $n_1$ and $n_2$.

3. For each permutation $i \in \{1, \ldots, M\}$, compute the permuted bias score:

$$\Delta^{(i)} = \hat{p}_1^{(i)} - \hat{p}_2^{(i)}$$

4. Estimate the two-tailed $p$-value:

$$p = \frac{1}{M} \sum_{i=1}^{M} \mathbb{I} \left( \left| \Delta^{(i)} \right| \geq \left| \Delta_{\text{obs}} \right| \right)$$

where $\mathbb{I}(\cdot)$ is the indicator function, and $M$ is the number of permutations (e.g., 10,000).

If $p < 0.05$, we consider the observed bias statistically significant.

## C.2 Rubric-based Evaluation

### C.2.1 Rubrics Interface and Diagnostic Evaluation

In the Rubric-based evaluation interface, annotators were required to assess each model's performance in isolation. For each benchmark dimension, the system iterated through all models and presented their dialogue outputs as audio. Alongside each audio sample, annotators were shown a natural language explanation of the specific capability being tested. A sample inferface used for Rubric-based evaluation is shown in Figure 14.

Each Rubrics task includes:

- A textual explanation of the evaluation objective

- A system prompt and user input (in audio form)

- A single model response (in audio form)

- A list of nine binary rubric criteria to be checked

This format encourages diagnostic evaluation by isolating each model's performance and aligning it explicitly with the desired communicative capability.

### C.2.2 Hierarchical Rubric Design

To support structured, interpretable, and reliable model evaluation, our rubric design follows a hierarchical three-level schema inspired by principles of educational assessment Suskie (2018). This hierarchy ensures both consistency across evaluations and adaptability to instance-specific capabilities while maintaining systematic coverage of all relevant evaluation aspects.

**Level 1: General Rubrics**   These rubrics are applicable across all tasks and dimensions, capturing broad qualities such as fluency, grammatical correctness, and relevance to the user prompt. General rubrics serve as a foundational layer to ensure basic communicative quality is met regardless of the evaluation focus, providing a consistent baseline for comparison across different evaluation scenarios.

**Level 2: Dimension-Specific Rubrics**   Each benchmark dimension—*Semantic Information*, *Paralinguistic Information*, and *Ambient Sound*—is associated with a dedicated rubric set targeting its core capabilities:

- **Semantic**: we focus on discourse coherence, contextual accuracy, and relevance to prior turns.

- **Paralinguistic**: we examine emotional clarity, intonation appropriateness, and disfluency handling.

- **Ambient**: we assess robustness to background noise, environment-aware response, and signal preservation.

These rubrics are curated by expert annotators and refined through pilot evaluations to ensure construct validity and dimension alignment.

**Level 3: Sample-Level Rubrics** At the most granular level, we use a large language model to generate contextualized rubrics specific to each dialogue instance. For every evaluation sample, the LLM receives the user-system exchange and a target evaluation goal (e.g., "emotional fidelity"), and generates a binary rubric such as: *"Does the response reflect the speaker's intended emotion of disappointment?"* To ensure the clarity and alignment of these automatically generated rubrics with the overall evaluation objective, each LLM-produced rubric is manually reviewed and revised if necessary by a trained annotator. This human-in-the-loop process ensures interpretability and minimizes ambiguity Hashemi et al. (2024).

**Annotation Guidelines and Review** All rubric reviews followed a standardized protocol to maintain quality and consistency. Rubrics were accepted only if their scope matched the designated dimension and avoided overlapping with general-purpose criteria. Ambiguous or overly subjective rubrics were flagged and rewritten for clarity, while rubrics with untestable or ill-posed binary conditions were discarded. Across the dataset, over **1350** sample-level rubrics were reviewed, with an acceptance rate of approximately **92%**, demonstrating the effectiveness of our quality control process.

# D   Experiment

This appendix provides supplementary information regarding the experimental implementation referenced in Section 6. It includes details on the computing infrastructure, models evaluated, and the protocols for human and LLM-based evaluation.

## D.1   Evaluation Setup

### D.1.1   Evaluation Models

- **GPT-4o Realtime**[3]: A multimodal model by OpenAI, supporting real-time S2S interaction with expressive prosody and advanced perception capabilities.

- **Doubao**[4]: A conversational AI model developed by ByteDance, integrated into a wide range of applications and known for its natural language interaction capabilities.

- **Kimi-Audio**[5]: An audio-capable model from Moonshot AI, specializing in long-context understanding and processing spoken dialogue.

- **MindGPT-4o-Audio**[6]: An in-car voice assistant developed by Li Auto, optimized for vehicle control, navigation, and entertainment through speech commands.

- **Moshi**[7]: A multimodal model from 01.AI, designed for seamless integration of text and speech processing in conversational applications.

- **GLM-4-Voice-9B**[8]: An open-source, end-to-end S2S model from Zhipu AI and Tsinghua University, optimized for bilingual (Chinese/English) multi-turn speech interaction.

- **Qwen-Omni-Turbo**[9]: A fully multimodal model from Alibaba Cloud capable of processing and generating audio, text, and images, supporting real-time dialogue.

- **Westlake-Omni**[10]: A multimodal conversational model from Westlake University, designed for prosodic and emotion-aware speech interaction.

- **VITA-Audio-Plus-Vanilla**[11]: An open-source multimodal model focused on integrating visual and audio information for speech-based tasks.

---

[3]Accessed via OpenAI API, May 2025. https://openai.com/index/hello-gpt-4o

[4]Accessed via the Doubao platform from ByteDance, May 2025. https://seed.bytedance.com/en/special/realtime_voice

[5]Accessed via the Kimi platform from Moonshot AI, May 2025. https://github.com/MoonshotAI/Kimi-Audio

[6]Accessed via Li Auto's in-car intelligent system, May 2025. https://www.lixiang.com/en/mega

[7]Accessed via 01.AI's platform, May 2025. https://github.com/kyutai-labs/moshi

[8]Accessed from the official repository, May 2025. https://github.com/THUDM/GLM-4

[9]Accessed from the official website, May 2025. https://github.com/QwenLM/Qwen2.5-Omni

[10]Accessed from the official repository, May 2025. https://github.com/xinchen-ai/Westlake-Omni

[11]Accessed from the official repository, May 2025. https://github.com/VITA-MLLM/VITA-Audio

- **AnyGPT**[12]: A multimodal model from Tencent ARC Lab that unifies text, speech, image, and music generation within a single framework.
- **MiniCPM-o 2.6**[13]: An open-source, efficient multimodal model capable of speech and image understanding and generation, developed by the OpenBMB community.
- **SpeechGPT 2.0-preview**[14]: An open-source large language model designed to follow complex speech-text instructions and solve various speech-related tasks.
- **Step-Audio-Chat**[15]: An open-source model designed for audio story generation, capable of creating coherent narratives from text prompts with corresponding sound effects and music.

### D.1.2 Computing Infrastructure

Experiments were conducted on two distinct server configurations:

**For Step-Audio-Chat:** Inference is performed on a server with two AMD EPYC 7742 64-Core CPUs, 1.0 TB of RAM, and four NVIDIA A100-SXM4 GPUs (80 GB VRAM each), running Ubuntu 22.04.4 LTS.

**For all other models:** Experiments are run on a server with an Intel Xeon Silver 4310 12-Core CPU, 251 GB of RAM, and a single NVIDIA GeForce RTX 4090 D GPU (24 GB VRAM), running Rocky Linux 8.10.

### D.1.3 Experiment on Microphone-based Input

To examine whether microphone-based input produces notable differences in output quality compared to direct audio file input, we design a controlled experiment focusing on models that support both modalities. This study is particularly relevant for models whose architectures are not publicly released and that do not provide API access—such as Doubao, Lixiang, and Moshi—where evaluation can only be performed through their desktop or web-based clients.

**Experimental Setup**    Three speech-to-speech models (GPT-4o Realtime, Qwen-Omni-Turbo, and GLM-4-Voice) are evaluated on 30 multi-turn dialogue samples (10 per dimension) covering all MTalk-Bench scenarios. Each dialogue is tested under two conditions: (1) direct audio file input via API, and (2) microphone-based input via real-time playback and recapture under controlled acoustic conditions. Three trained annotators independently assess each response pair following MTalk-Bench guidelines.

**Results and Analysis**    The results in Table 8 show that across all models, the majority of comparisons indicate equivalent quality, with Qwen-Omni-Turbo achieving the highest consistency (**73.3%** equivalent outputs), followed by GLM-4-Voice (**56.7%**) and GPT-4o (**50.0%**). GPT-4o shows the largest proportion of cases (**33.3%**) favoring direct audio input, suggesting higher sensitivity to microphone-based capture. GLM-4-Voice exhibits **30.0%** of cases where microphone input yields better results, potentially due to robustness in speech processing. These findings suggest that while microphone-mediated evaluation is a valid alternative for models without API access, subtle model-specific variations exist and should be considered when interpreting performance differences.

| Model | Same quality | Direct audio file input better | Microphone-based input better |
|---|---|---|---|
| GLM-4-Voice | **56.7%** (17) | 13.3% (4) | 30.0% (9) |
| GPT-4o | **50.0%** (15) | 33.3% (10) | 16.7% (5) |
| Qwen-Omni-Turbo | **73.3%** (22) | 26.7% (8) | – |

Table 8: Comparison of output quality between direct audio file input and microphone-based input for three S2S models (30 samples per model).

---

### D.2 Turn-level Analysis Methodology

To investigate performance degradation across dialogue turns, we conduct a comprehensive turn-level analysis examining both response quality and content density as conversations progress. This analysis provides insights into how different models maintain coherence and informativeness in extended multi-turn interactions.

**Response Quality Calculation**   Response quality is measured using adapted Rubric-based evaluation scores for truncated dialogues. We generate specialized rubrics tailored to incomplete first and second turns, following the same prompt template used in our standard Rubric-based evaluation with GPT-4o Realtime. For each model and each turn position (T1, T2, T3), we apply these turn-specific rubrics to assess the quality of responses up to that point in the conversation.

**Content Density Calculation**   Content density measures the proportion of essential information in model responses by evaluating clause-level necessity. For each dialogue truncated at turn T1, T2, or T3, we employ GPT-4o Realtime to assess whether each complete clause in the final response turn can be removed without losing critical information. The model evaluates each clause for its contribution to the overall dialogue goal and context. Content density is calculated as the ratio of word count in essential (non-removable) clauses to the total word count in the response, expressed as a percentage. This metric captures how efficiently models convey information without unnecessary verbosity or redundancy.

### D.3 Evaluator Protocols

We employ both human and LLM evaluators to ensure comprehensive assessment while providing baseline comparisons for automated evaluation methods.

#### D.3.1 Human Evaluation Protocol

To ensure the validity and scalability of our human evaluation, we deploy the evaluation interface through two parallel channels: (1) a link distributed to internal annotators recruited via university mailing lists and research communities, and (2) the Amazon Mechanical Turk (MTurk) platform. Both groups of annotators are provided with **identical evaluation instructions**, interfaces, and task formats to ensure consistency across the two channels.

For the **University annotators**, all participants are undergraduate students at the time of participation. Recruitment is conducted via online surveys circulated through campus mailing lists. Annotators are required to self-report their English listening proficiency by submitting official score records. Specifically, only participants with an IELTS listening score of 6.5 or above, or a TOEFL listening score of 25 or above, are eligible to proceed. Prior to annotation, each qualified annotator receives standardized training that includes detailed task instructions and illustrative examples to ensure consistent and accurate evaluations. Compensation is provided based on the volume of valid annotations completed, with a rate of 1.5 to 2 RMB per valid evaluation item.

For the **MTurk annotators**, we restrict participation to workers holding the *Masters Qualification* on Amazon Mechanical Turk, with a HIT Approval Rate greater than 65% across all requesters' HITs. Additionally, annotators must be located in one of the following English-speaking countries: Australia (AU), Canada (CA), Ireland (IE), New Zealand (NZ), United Kingdom (GB), or United States (US). Tasks are posted through the official MTurk platform and compensation is issued at a rate of $0.40 to $0.50 per group of valid annotations.

**Quality Filtering Process**   A multi-stage process is implemented to ensure the reliability of human-provided data:

- **Minimum Engagement:** The evaluation interface enforces a minimum time on each page equal to the total audio playback duration.

- **Consistency Filtering:** A custom script flags annotations that conflict with evaluation instructions or focus on irrelevant attributes (e.g., accent naturalness in unrelated tasks).

- **Manual Review:** Experts review remaining annotations to remove spam or low-effort responses (e.g., repeated rationale texts).

### D.3.2 LLM-as-Judge Protocol

**Raw Audio Evaluation** The prompts shown to LLM judges are designed to mirror the human evaluation tasks and criteria.

- **Pairwise Arena Judgment:** The LLM receives the input audio and the answer audios of two model responses. It is prompted to select the better response based on specific capability (see prompt template in Figure 15).
- **Absolute Rubric Scoring:** The LLM judges receive the input and answer audios, along with a detailed rubric. Its task is to score the response against each criterion and provide a structured justification, mirroring the human process (see prompt templates in Figures 16, 17, and 18.

**Transcribed Text Evaluation** This evaluation modality follows the same structure as the audio protocol but uses transcribed text instead of audio. The LLM receives transcripts of both the input audio and model responses for pairwise and Rubric-based scoring tasks, along with any available paralinguistic or ambient sound tags.

## E  Evaluation Result

This section provides additional evaluation results to complement the main text. Figure 11 provides a capability-wise comparison of S2S models, illustrating their strengths and weaknesses across nine communicative functions as judged by both humans and LLMs. We present detailed performance results for both raw speech and transcript-based evaluations. Tables 9 and 12 summarize the analysis of speech modalities, while Tables 10 and 11 cover the transcript analysis. All tables report on performance under both Arena and Rubric-based settings across three dimensions. Figure 12 further analyzes the consistency of Elo ratings with raw win-rate statistics.

### E.1  Detailed Analysis of Experiment

**Evaluation of S2S LLMs across Different Capabilities** Figure 11 presents a comparative analysis of rubrics performance across nine fine-grained capabilities, evaluated independently by human judges (left) and the GPT-4o Realtime model (right).

Under human evaluation, GPT-4o Realtime demonstrates strong performance in *Core Comprehension and Memory* and *Ambient Sound Understanding*. The Human baseline also scores highly in *Security Assessment*. Qwen-Omni-Turbo shows notable strength in *Paralinguistic Comprehension* and *Interaction Strategy and Intelligence*. In contrast, AnyGPT consistently underperforms across all capabilities.

In comparison, GPT-4o Realtime-as-judge produces more uniform scores across models and tasks. While some trends align with human judgment (e.g., *Core Comprehension*), key dimensions like *Paralinguistic Generation* and *Ambient Sound Understanding* show notable discrepancies—highlighting current limitations of LLM-based evaluation, especially in non-textual domains.

**Correlation Between Elo and Win-rate** Figure 12 plots Elo score differences against empirical win-rate differences across all system pairs. The resulting Spearman correlation ($\rho = 0.838$) indicates a strong monotonic relationship: Elo ratings faithfully summarize pairwise preferences while smoothing over sampling noise. To ensure fairness, Elo was recomputed on the merged dataset, respecting temporal order when available and applying random shuffling within unordered batches. The high correlation validates Elo as an effective system-level indicator for model comparison.

### E.2  Evaluation Consistency Analysis

**Human Evaluator Consistency** Table 12 separates human judgments into two groups: public link and MTurk workers. The left sub-table shows Elo scores, and the right sub-table reports Rubric-based scores. The overall rankings across the two sources remain correlated, underscoring the robustness of our conclusions. Small differences emerge in paralinguistic and ambient evaluations, likely reflecting annotators' attention span, listening conditions, or domain familiarity. Nevertheless, the consistency at the top and bottom tiers supports the reliability of aggregated human assessments.

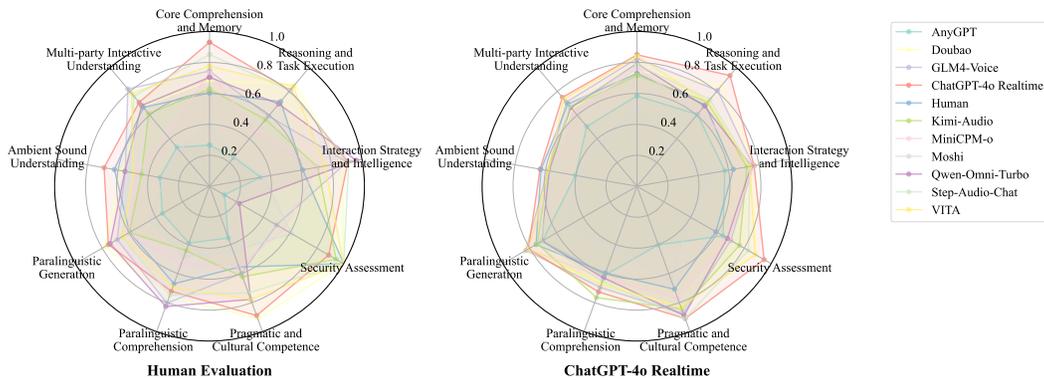

**Human Evaluation**  **ChatGPT-4o Realtime**

Figure 11: Rubric-based results in 9 capability dimensions.

| S2S Models | Human | | | | GPT-4o Realtime | | | | Gemini-2.5-pro | | | | Qwen-Omni-Turbo | | | |
|---|---|---|---|---|---|---|---|---|---|---|---|---|---|---|---|---|
| | Sem. | Para. | Ambi. | Ovrl.↑ | Sem. | Para. | Ambi. | Ovrl. | Sem. | Para. | Ambi. | Ovrl. | Sem. | Para. | Ambi. | Ovrl. |
| *Arena-style Evaluation* | | | | | | | | | | | | | | | | |
| *Closed-source Models* | | | | | | | | | | | | | | | | |
| GPT-4o Realtime | 1052(71) | 1038(81) | 1026(83) | 1039(259) | **1043(75)** | 1008(66) | **1030(60)** | 1027(229) | 1019(57) | 1021(76) | **1029(41)** | 1023(170) | 1022(119) | 1003(79) | 1022(120) | 1016(318) |
| Doubao | 1022(52) | 1039(70) | 1047(46) | 1036(306) | 1017(48) | 1024(66) | 1007(68) | 1016(152) | **1028(55)** | **1031(36)** | 1013(60) | **1024(113)** | 1029(38) | 1010(59) | 1007(98) | 1007(145) |
| Qwen-Omni-Turbo | 1004(63) | 1044(71) | **1054(87)** | 1034(241) | 993(135) | 1023(129) | 995(97) | 1004(267) | 994(90) | 989(111) | 975(112) | 986(322) | 1023(56) | 1012(76) | 1004(91) | 1013(207) |
| *Open-source Models* | | | | | | | | | | | | | | | | |
| Step-Audio-Chat | **1072(84)** | **1052(79)** | 1038(80) | 1054(213) | 1028(92) | 1006(92) | 1002(92) | **1027(321)** | 994(118) | 990(97) | 992(119) | 992(334) | **1046(85)** | 1021(76) | **1051(88)** | **1035(281)** |
| VITA-Audio-Plus-Vanilla | 1009(116) | 973(83) | 1009(91) | 997(292) | 1018(111) | 1000(84) | 1014(44) | 1014(281) | 1019(116) | 1021(114) | 1016(109) | 1019(341) | 1022(88) | 1000(68) | 1012(55) | 1011(187) |
| GLM-4-Voice | 994(148) | 1006(99) | 988(76) | 996(253) | 1007(111) | 1011(90) | 1010(80) | 1012(331) | 1019(114) | 1021(114) | 1016(109) | 1019(341) | 1000(64) | 1012(63) | 1000(64) | 1011(187) |
| Kimi-Audio | 999(76) | 995(55) | 979(60) | 991(199) | 988(85) | 1007(83) | 1002(79) | **1075(116)** | 1007(90) | 1031(101) | 1014(69) | 1018(260) | 978(46) | 973(26) | 981(62) | 987(230) |
| Westlake-Omni | 986(7) | 989(65) | 992(8) | 989(21) | 969(48) | 937(76) | 979(45) | 962(171) | 980(44) | 975(91) | 987(58) | 981(166) | 984(46) | 973(26) | 983(44) | 977(71) |
| SpeechGPT 2.0-preview | 989(69) | 979(111) | 1000(81) | 989(23) | 978(33) | 938(47) | 966(51) | 961(109) | 971(57) | 983(31) | 971(91) | 975(227) | 977(45) | 983(44) | 980(32) | 980(136) |
| MiniCPM-o 2.6 | 997(117) | 993(87) | 975(82) | 988(386) | 982(73) | 985(105) | 994(75) | 987(253) | 979(57) | 988(83) | 990(117) | 985(267) | 986(132) | 1002(115) | 991(42) | 998(364) |
| MindGPT-4o-Audio | 986(9) | 998(83) | 981(20) | 988(36) | 1018(111) | 1031(48) | 1024(46) | 1024(167) | 1003(51) | 1018(57) | 1028(70) | 1022(199) | 997(123) | 991(42) | 1004(62) | 997(207) |
| Moshi | 959(47) | 975(88) | 952(55) | 962(170) | 990(49) | 996(57) | 996(46) | 994(182) | 1001(53) | 1004(51) | 997(55) | 1001(159) | 985(60) | 992(64) | 987(68) | 988(188) |
| AnyGPT | 942(51) | 945(67) | 928(61) | 938(179) | 982(51) | 948(67) | 989(55) | 973(173) | 985(60) | 983(44) | 973(173) | 973(173) | 992(38) | 1000(53) | 966(70) | 988(181) |
| Human | 990(129) | 975(101) | 1031(77) | 999(156) | 959(104) | 941(124) | 978(114) | 959(342) | 972(100) | 981(91) | 1003(74) | 985(267) | 983(91) | 966(91) | 967(97) | 972(279) |
| *Rubric-based Evaluation* | | | | | | | | | | | | | | | | |
| *Closed-source Models* | | | | | | | | | | | | | | | | |
| GPT-4o Realtime | **88.59(90)** | 73.75(50) | **69.73(49)** | **77.38(149)** | **88.06(96)** | 76.01(96) | 67.80(60) | 77.33(270) | **73.76(90)** | **64.34(90)** | **70.31(90)** | **69.50(270)** | 84.85(88) | 80.3(57) | **86.51(97)** | **84.01(264)** |
| Doubao | 82.54(50) | 77.00(50) | 64.53(47) | 73.69(145) | 81.97(90) | 73.61(90) | 60.50(90) | 72.06(270) | 70.02(90) | 60.31(90) | 60.63(90) | 63.69(270) | 82.73(83) | 80.28(86) | 82.07(82) | 81.71(251) |
| Qwen-Omni-Turbo | 76.08(48) | **78.83(48)** | 60.05(47) | 71.82(143) | 74.88(90) | 66.29(90) | 60.75(90) | 67.34(270) | 54.35(90) | 47.59(90) | 56.73(90) | 52.91(270) | 82.83(83) | 80.71(88) | 84.48(89) | 82.72(265) |
| *Open-source Models* | | | | | | | | | | | | | | | | |
| Step-Audio-Chat | 85.50(47) | 70.20(49) | 59.55(48) | 71.86(145) | 79.23(90) | 67.80(90) | 59.87(90) | 69.01(270) | 62.81(93) | 55.64(90) | 56.23(90) | 58.25(270) | 81.39(83) | 79.53(82) | 84.04(85) | 81.35(281) |
| GLM-4-Voice | 72.41(50) | 75.45(49) | 64.33(48) | 70.81(147) | 79.10(90) | 74.87(90) | 63.02(90) | 72.35(270) | 57.09(90) | 57.43(90) | 57.48(90) | 57.33(270) | 82.04(86) | 80.46(83) | 82.94(87) | 81.86(256) |
| VITA-Audio-Plus-Vanilla | 79.56(50) | 57.46(50) | 59.22(48) | 68.70(146) | 79.23(90) | 72.73(90) | 63.77(90) | 71.94(270) | 55.89(90) | 57.11(90) | 57.50(90) | 56.83(270) | 80.81(88) | 80.54(89) | 85.20(88) | 82.20(265) |
| MiniCPM-o 2.6 | 78.36(50) | 57.11(47) | 46.04(48) | 55.74(141) | 71.64(90) | 70.45(90) | 58.13(90) | 66.75(270) | 49.63(90) | 51.07(90) | 49.69(90) | 50.13(270) | 75.28(89) | 75.67(87) | 78.51(88) | 76.50(264) |
| Kimi-Audio | 65.56(48) | 50.10(46) | 50.00(51) | 55.17(145) | 74.86(82) | 75.63(91) | 61.73(84) | 70.66(247) | 55.33(92) | 60.62(85) | 58.63(84) | 58.17(247) | 78.59(82) | 79.49(81) | 76.01(84) | 77.98(247) |
| Moshi | 35.05(46) | 46.58(49) | 29.58(48) | 37.26(143) | 46.39(90) | 47.15(90) | 37.74(90) | 43.76(270) | 32.15(90) | 28.68(90) | 30.18(90) | 31.62(270) | 58.99(83) | 61.62(86) | 58.99(90) | 60.55(265) |
| AnyGPT | 28.57(49) | 37.39(49) | 32.47(48) | 32.76(146) | 56.38(90) | 64.90(90) | 43.90(90) | 55.05(270) | 15.30(90) | 11.94(90) | 7.04(90) | 11.44(270) | 37.20(89) | 38.87(88) | 44.03(90) | 40.06(267) |
| MindGPT-4o-Audio | 35.42(50) | 32.67(49) | 32.74(48) | 33.63(147) | 84.08(90) | **79.55(90)** | **71.07(90)** | **78.25(270)** | 67.16(90) | 63.50(90) | 63.77(90) | 64.82(270) | 83.27(89) | **81.85(90)** | 83.72(90) | 82.97(268) |
| SpeechGPT 2.0-preview | 2.90(50) | 8.35(50) | 6.49(49) | 5.96(149) | 18.91(90) | 16.79(90) | 17.82(90) | 17.82(270) | 16.49(90) | 16.16(90) | 16.07(90) | 16.18(270) | 8.40(89) | 13.62(90) | 8.40(89) | 11.18(268) |
| Westlake-Omni | 3.11(49) | 6.47(50) | 5.94(49) | 5.18(148) | 23.76(90) | 25.51(90) | 20.00(90) | 23.17(270) | 14.83(90) | 19.79(90) | 13.46(90) | 15.60(270) | | | | |
| Human | 65.25(50) | 66.67(50) | 69.06(47) | 66.95(147) | 71.14(90) | 65.91(90) | 61.51(90) | 66.21(270) | 47.64(90) | 54.27(90) | 58.87(90) | 53.57(270) | 72.73(90) | 73.08(90) | 74.47(90) | 71.32(270) |

Table 9: Combined Evaluation Results: Arena-style ELO scores are rounded to the nearest integer, with Semantic and Ambient columns swapped. Rows are sorted by the Human Overall (Ovrl.↑) scores in descending order. ↑ indicates rows are ranked by Human Overall scores (high to low). Each score is accompanied by a subscript in parentheses (e.g., $(50)$), indicating the number of votes on which the score is based.

| S2S Models | GPT-4o Realtime | | | | Claude Sonnet 4 | | | | Claude Sonnet 4 Thinking | | | | DeepSeek R1 | | | |
|---|---|---|---|---|---|---|---|---|---|---|---|---|---|---|---|---|
| | *Sem.* | *Para.* | *Ambi.* | *Orcl.↑* | *Sem.* | *Para.* | *Ambi.* | *Orcl.* | *Sem.* | *Para.* | *Ambi.* | *Orcl.* | *Sem.* | *Para.* | *Ambi.* | *Orcl.* |
| *Arena-style Evaluation* | | | | | | | | | | | | | | | | |
| *Closed-source Models* | | | | | | | | | | | | | | | | |
| Doubao | 1025$_{(35)}$ | **1027$_{(24)}$** | **1022$_{(41)}$** | **1025$_{(100)}$** | 1022$_{(47)}$ | 1021$_{(29)}$ | 1023$_{(43)}$ | 1023$_{(106)}$ | 1021$_{(45)}$ | **1026$_{(25)}$** | 1016$_{(46)}$ | 1021$_{(100)}$ | 1011$_{(60)}$ | **1022$_{(31)}$** | 1013$_{(57)}$ | 1016$_{(120)}$ |
| GPT-4o Realtime | **1035$_{(36)}$** | 1020$_{(34)}$ | 1019$_{(28)}$ | 1025$_{(92)}$ | **1026$_{(31)}$** | **1024$_{(38)}$** | **1024$_{(34)}$** | **1024$_{(95)}$** | **1026$_{(47)}$** | 1029$_{(29)}$ | **1029$_{(29)}$** | **1026$_{(13)}$** | **1027$_{(26)}$** | 1017$_{(43)}$ | **1028$_{(44)}$** | **1024$_{(85)}$** |
| Qwen-Omni-Turbo | 1002$_{(62)}$ | 1006$_{(56)}$ | 1009$_{(51)}$ | 1006$_{(172)}$ | 1011$_{(60)}$ | 1013$_{(63)}$ | 1005$_{(53)}$ | 1008$_{(155)}$ | 1011$_{(74)}$ | 1015$_{(48)}$ | 1021$_{(49)}$ | 1016$_{(135)}$ | 1007$_{(51)}$ | 1005$_{(47)}$ | 997$_{(57)}$ | 1000$_{(155)}$ |
| *Open-source Models* | | | | | | | | | | | | | | | | |
| Step-Audio-Chat | 1016$_{(54)}$ | 1019$_{(62)}$ | 1005$_{(48)}$ | 1013$_{(164)}$ | 1018$_{(65)}$ | 998$_{(47)}$ | 1010$_{(70)}$ | 1009$_{(182)}$ | 1008$_{(48)}$ | 1000$_{(47)}$ | 999$_{(77)}$ | 1005$_{(172)}$ | 1013$_{(60)}$ | 1007$_{(55)}$ | 1005$_{(45)}$ | 1008$_{(168)}$ |
| GLM-4-Voice | 1007$_{(54)}$ | 1017$_{(57)}$ | 1007$_{(36)}$ | 1010$_{(149)}$ | 1011$_{(60)}$ | 1011$_{(69)}$ | 1004$_{(68)}$ | 1008$_{(197)}$ | 1013$_{(63)}$ | 1019$_{(64)}$ | 1005$_{(60)}$ | 1015$_{(191)}$ | 1018$_{(60)}$ | 1009$_{(38)}$ | 1013$_{(48)}$ | 1015$_{(127)}$ |
| VITA-Audio-Plus-Vanilla | 1007$_{(52)}$ | 1000$_{(61)}$ | 1014$_{(53)}$ | 1007$_{(145)}$ | 1005$_{(49)}$ | 1012$_{(78)}$ | 1022$_{(65)}$ | 1013$_{(146)}$ | 1014$_{(53)}$ | 1019$_{(64)}$ | 1005$_{(66)}$ | 1018$_{(141)}$ | 1021$_{(29)}$ | 1001$_{(60)}$ | 1018$_{(47)}$ | 1013$_{(119)}$ |
| MiniCPM 2.6 | 996$_{(46)}$ | 996$_{(54)}$ | 1007$_{(52)}$ | 1000$_{(152)}$ | 998$_{(59)}$ | 997$_{(69)}$ | 995$_{(71)}$ | 997$_{(199)}$ | 991$_{(45)}$ | 1000$_{(58)}$ | 1010$_{(53)}$ | 1005$_{(156)}$ | 994$_{(55)}$ | 998$_{(48)}$ | 1003$_{(56)}$ | 998$_{(159)}$ |
| Moshi | 995$_{(50)}$ | 999$_{(20)}$ | 994$_{(35)}$ | 996$_{(105)}$ | 995$_{(53)}$ | 991$_{(51)}$ | 988$_{(38)}$ | 991$_{(122)}$ | 996$_{(36)}$ | 997$_{(49)}$ | 995$_{(52)}$ | 996$_{(139)}$ | 996$_{(49)}$ | 992$_{(42)}$ | 997$_{(53)}$ | 995$_{(53)}$ |
| Kimi-Audio | 986$_{(41)}$ | 993$_{(52)}$ | 996$_{(46)}$ | 992$_{(139)}$ | 999$_{(41)}$ | 1009$_{(60)}$ | 1005$_{(64)}$ | 1004$_{(165)}$ | 1003$_{(52)}$ | 994$_{(41)}$ | 1003$_{(44)}$ | 1000$_{(137)}$ | 998$_{(50)}$ | 995$_{(61)}$ | 999$_{(50)}$ | 1000$_{(165)}$ |
| MindGPT-4o-Audio | 987$_{(27)}$ | 978$_{(21)}$ | 988$_{(30)}$ | 984$_{(78)}$ | 977$_{(36)}$ | 978$_{(21)}$ | 975$_{(27)}$ | 977$_{(84)}$ | 981$_{(32)}$ | 983$_{(34)}$ | 984$_{(26)}$ | 980$_{(41)}$ | 984$_{(28)}$ | 980$_{(41)}$ | 976$_{(30)}$ | 982$_{(107)}$ |
| AnyGPT | 980$_{(40)}$ | 986$_{(48)}$ | 987$_{(35)}$ | 982$_{(111)}$ | 989$_{(45)}$ | 986$_{(41)}$ | 989$_{(45)}$ | 990$_{(117)}$ | 990$_{(45)}$ | 985$_{(42)}$ | 991$_{(47)}$ | 991$_{(147)}$ | 999$_{(60)}$ | 996$_{(53)}$ | 994$_{(30)}$ | 994$_{(130)}$ |
| Westlake-Omni | 979$_{(17)}$ | 974$_{(32)}$ | 982$_{(15)}$ | 978$_{(53)}$ | 974$_{(46)}$ | 974$_{(15)}$ | 980$_{(10)}$ | 978$_{(70)}$ | 978$_{(25)}$ | 970$_{(21)}$ | 982$_{(17)}$ | 973$_{(26)}$ | 975$_{(55)}$ | 976$_{(12)}$ | 980$_{(18)}$ | 978$_{(66)}$ |
| SpeechGPT 2.0-preview | 973$_{(18)}$ | 979$_{(25)}$ | 978$_{(19)}$ | 976$_{(62)}$ | 974$_{(21)}$ | 974$_{(15)}$ | 978$_{(11)}$ | 976$_{(63)}$ | 975$_{(43)}$ | 982$_{(21)}$ | 972$_{(20)}$ | 976$_{(84)}$ | 977$_{(36)}$ | 976$_{(12)}$ | 980$_{(18)}$ | 978$_{(66)}$ |
| Human | 1000$_{(31)}$ | 991$_{(65)}$ | 979$_{(63)}$ | 990$_{(169)}$ | 995$_{(48)}$ | 1002$_{(48)}$ | 1001$_{(44)}$ | 999$_{(144)}$ | 991$_{(60)}$ | 963$_{(59)}$ | 994$_{(46)}$ | 982$_{(168)}$ | 977$_{(50)}$ | 1014$_{(58)}$ | 976$_{(55)}$ | 989$_{(163)}$ |
| *Rubric-based Evaluation* | | | | | | | | | | | | | | | | |
| *Closed-source Models* | | | | | | | | | | | | | | | | |
| GPT-4o Realtime | **78.36$_{(90)}$** | **65.91$_{(90)}$** | **66.79$_{(90)}$** | **70.39$_{(270)}$** | **66.17$_{(90)}$** | **47.60$_{(90)}$** | **58.99$_{(90)}$** | **57.63$_{(270)}$** | **71.14$_{(90)}$** | **48.86$_{(90)}$** | **62.77$_{(90)}$** | **60.98$_{(270)}$** | **68.78$_{(90)}$** | **51.89$_{(90)}$** | **59.62$_{(90)}$** | **60.14$_{(270)}$** |
| Doubao | 73.75$_{(90)}$ | 63.70$_{(90)}$ | 56.68$_{(88)}$ | 65.50$_{(268)}$ | 65.55$_{(90)}$ | 43.69$_{(90)}$ | 47.81$_{(88)}$ | 52.44$_{(268)}$ | 68.28$_{(90)}$ | 48.29$_{(90)}$ | 52.00$_{(88)}$ | 56.97$_{(268)}$ | 66.54$_{(90)}$ | 49.24$_{(90)}$ | 50.51$_{(90)}$ | 55.52$_{(268)}$ |
| Qwen-Omni-Turbo | 64.30$_{(90)}$ | 52.53$_{(90)}$ | 52.70$_{(90)}$ | 56.55$_{(270)}$ | 51.87$_{(90)}$ | 38.01$_{(90)}$ | 46.54$_{(90)}$ | 45.91$_{(270)}$ | 56.97$_{(90)}$ | 40.83$_{(90)}$ | 50.31$_{(90)}$ | 49.41$_{(270)}$ | 62.58$_{(90)}$ | 42.05$_{(90)}$ | 45.41$_{(90)}$ | 47.34$_{(270)}$ |
| *Open-source Models* | | | | | | | | | | | | | | | | |
| VITA-Audio-Plus-Vanilla | 69.65$_{(90)}$ | 60.48$_{(90)}$ | 57.76$_{(90)}$ | 62.68$_{(269)}$ | 54.98$_{(90)}$ | 41.92$_{(90)}$ | 49.36$_{(90)}$ | 48.78$_{(269)}$ | 58.96$_{(90)}$ | 47.85$_{(90)}$ | 52.80$_{(90)}$ | 53.23$_{(269)}$ | 58.21$_{(90)}$ | 49.12$_{(90)}$ | 49.87$_{(90)}$ | 52.43$_{(269)}$ |
| GLM-4-Voice | 63.56$_{(90)}$ | 57.45$_{(90)}$ | 52.58$_{(90)}$ | 57.88$_{(270)}$ | 49.25$_{(90)}$ | 41.41$_{(90)}$ | 41.76$_{(90)}$ | 44.11$_{(270)}$ | 52.86$_{(90)}$ | 44.32$_{(90)}$ | 47.17$_{(90)}$ | 48.14$_{(270)}$ | 55.85$_{(90)}$ | 44.82$_{(90)}$ | 43.90$_{(90)}$ | 48.22$_{(270)}$ |
| Step-Audio-Chat | 62.56$_{(90)}$ | 54.92$_{(90)}$ | 48.95$_{(90)}$ | 56.96$_{(270)}$ | 59.83$_{(90)}$ | 42.42$_{(90)}$ | 44.03$_{(90)}$ | 48.81$_{(270)}$ | 53.34$_{(90)}$ | 44.95$_{(90)}$ | 47.30$_{(90)}$ | 50.56$_{(270)}$ | 57.95$_{(90)}$ | 44.82$_{(90)}$ | 45.79$_{(90)}$ | 49.44$_{(270)}$ |
| MiniCPM 2.6 | 58.99$_{(90)}$ | 48.79$_{(90)}$ | 42.88$_{(90)}$ | 50.25$_{(270)}$ | 48.81$_{(90)}$ | 37.68$_{(90)}$ | 40.71$_{(90)}$ | 42.43$_{(267)}$ | 52.33$_{(90)}$ | 39.59$_{(90)}$ | 44.78$_{(90)}$ | 45.60$_{(267)}$ | 51.82$_{(90)}$ | 40.61$_{(90)}$ | 41.22$_{(90)}$ | 44.57$_{(267)}$ |
| Kimi-Audio | 53.18$_{(90)}$ | 46.05$_{(90)}$ | 45.25$_{(87)}$ | 48.18$_{(264)}$ | 46.56$_{(90)}$ | 34.57$_{(90)}$ | 35.09$_{(90)}$ | 38.76$_{(265)}$ | 50.38$_{(90)}$ | 40.23$_{(90)}$ | 40.35$_{(90)}$ | 43.67$_{(265)}$ | 47.71$_{(90)}$ | 37.70$_{(90)}$ | 37.66$_{(90)}$ | 41.06$_{(265)}$ |
| SpeechGPT 2.0-preview | 66.54$_{(90)}$ | 22.22$_{(90)}$ | 28.29$_{(90)}$ | 39.30$_{(266)}$ | 0.48$_{(90)}$ | 0.00$_{(90)}$ | 0.00$_{(90)}$ | 0.10$_{(270)}$ | 1.99$_{(90)}$ | 0.13$_{(90)}$ | 0.02$_{(90)}$ | 0.92$_{(270)}$ | 28.96$_{(90)}$ | 7.58$_{(90)}$ | 4.40$_{(90)}$ | 13.51$_{(270)}$ |
| Westlake-Omni | 66.29$_{(90)}$ | 23.80$_{(90)}$ | 26.99$_{(90)}$ | 39.26$_{(268)}$ | 0.50$_{(90)}$ | 0.00$_{(90)}$ | 0.00$_{(90)}$ | 0.10$_{(226)}$ | 3.05$_{(90)}$ | 0.06$_{(90)}$ | 1.09$_{(276)}$ | | 31.59$_{(90)}$ | 6.00$_{(90)}$ | 2.64$_{(90)}$ | 13.51$_{(270)}$ |
| Moshi | 33.96$_{(90)}$ | 27.84$_{(90)}$ | 24.81$_{(88)}$ | 28.92$_{(267)}$ | 30.47$_{(90)}$ | 22.35$_{(90)}$ | 23.01$_{(88)}$ | 25.33$_{(267)}$ | 33.33$_{(90)}$ | 26.95$_{(90)}$ | 29.95$_{(90)}$ | 30.11$_{(267)}$ | 36.94$_{(90)}$ | 29.12$_{(90)}$ | 26.92$_{(89)}$ | 30.82$_{(267)}$ |
| MindGPT-4o-Audio | 38.06$_{(90)}$ | 17.05$_{(90)}$ | 23.55$_{(90)}$ | 26.30$_{(268)}$ | 6.25$_{(90)}$ | 4.72$_{(90)}$ | 5.89$_{(90)}$ | 5.29$_{(268)}$ | 5.72$_{(90)}$ | 5.05$_{(90)}$ | 6.62$_{(90)}$ | 5.79$_{(268)}$ | 29.06$_{(90)}$ | 7.49$_{(90)}$ | 8.91$_{(90)}$ | 15.58$_{(268)}$ |
| AnyGPT | 18.24$_{(90)}$ | 14.73$_{(88)}$ | 13.21$_{(90)}$ | 15.40$_{(267)}$ | 19.75$_{(89)}$ | 11.89$_{(88)}$ | 13.71$_{(90)}$ | 15.14$_{(267)}$ | 19.87$_{(89)}$ | 15.12$_{(88)}$ | 18.36$_{(90)}$ | 17.81$_{(267)}$ | 26.92$_{(89)}$ | 17.44$_{(90)}$ | 19.50$_{(90)}$ | 21.32$_{(267)}$ |
| Human | 52.61$_{(90)}$ | 46.59$_{(90)}$ | 47.55$_{(90)}$ | 48.93$_{(270)}$ | 44.90$_{(90)}$ | 34.85$_{(90)}$ | 48.81$_{(90)}$ | 42.87$_{(270)}$ | 47.76$_{(90)}$ | 39.70$_{(90)}$ | 50.82$_{(90)}$ | 46.11$_{(270)}$ | 42.91$_{(90)}$ | 40.28$_{(90)}$ | 43.14$_{(90)}$ | 42.12$_{(270)}$ |

| S2S Models | Gemini 2.5 Pro | | | | Kimi K2 | | | | Grok 4 | | | | OpenAI o3 | | | |
|---|---|---|---|---|---|---|---|---|---|---|---|---|---|---|---|---|
| | *Sem.* | *Para.* | *Ambi.* | *Orcl.↑* | *Sem.* | *Para.* | *Ambi.* | *Orcl.* | *Sem.* | *Para.* | *Ambi.* | *Orcl.* | *Sem.* | *Para.* | *Ambi.* | *Orcl.* |
| *Arena-style Evaluation* | | | | | | | | | | | | | | | | |
| *Closed-source Models* | | | | | | | | | | | | | | | | |
| Doubao | **1025$_{(27)}$** | **1030$_{(33)}$** | 1020$_{(34)}$ | **1025$_{(84)}$** | 1018$_{(35)}$ | 1012$_{(30)}$ | 1019$_{(20)}$ | 1017$_{(49)}$ | 1018$_{(35)}$ | **1027$_{(26)}$** | 1014$_{(37)}$ | **1020$_{(88)}$** | 1021$_{(32)}$ | **1024$_{(29)}$** | 1011$_{(34)}$ | **1018$_{(84)}$** |
| GPT-4o Realtime | 1019$_{(24)}$ | 1020$_{(30)}$ | **1025$_{(28)}$** | 1022$_{(80)}$ | **1027$_{(28)}$** | **1029$_{(39)}$** | **1022$_{(45)}$** | **1026$_{(72)}$** | 1018$_{(39)}$ | 1020$_{(30)}$ | **1016$_{(24)}$** | 1018$_{(80)}$ | 1014$_{(23)}$ | 1018$_{(31)}$ | **1022$_{(41)}$** | 1018$_{(49)}$ |
| Qwen-Omni-Turbo | 1009$_{(27)}$ | 1014$_{(49)}$ | 998$_{(51)}$ | 1007$_{(127)}$ | 1010$_{(51)}$ | 1000$_{(42)}$ | 994$_{(49)}$ | 1001$_{(142)}$ | 1010$_{(51)}$ | 1000$_{(42)}$ | 994$_{(49)}$ | 1001$_{(142)}$ | 1005$_{(41)}$ | 1004$_{(46)}$ | 1015$_{(52)}$ | 1010$_{(135)}$ |
| *Open-source Models* | | | | | | | | | | | | | | | | |
| Step-Audio-Chat | 1020$_{(54)}$ | 1009$_{(41)}$ | 1008$_{(63)}$ | 1012$_{(158)}$ | 1010$_{(45)}$ | 1018$_{(43)}$ | 1012$_{(140)}$ | | 1025$_{(37)}$ | 1001$_{(40)}$ | 993$_{(62)}$ | 1006$_{(138)}$ | **1027$_{(52)}$** | 1008$_{(55)}$ | 1002$_{(51)}$ | 1013$_{(148)}$ |
| Kimi-Audio | 1004$_{(30)}$ | 1004$_{(54)}$ | 1014$_{(53)}$ | 1007$_{(155)}$ | 1017$_{(50)}$ | 1006$_{(60)}$ | 1007$_{(63)}$ | 1010$_{(163)}$ | 997$_{(47)}$ | 1020$_{(51)}$ | 1004$_{(60)}$ | 1007$_{(158)}$ | 996$_{(50)}$ | 1008$_{(52)}$ | 1013$_{(61)}$ | 1008$_{(150)}$ |
| GLM-4-Voice | 1006$_{(65)}$ | 1008$_{(55)}$ | 996$_{(42)}$ | 1003$_{(145)}$ | 1006$_{(51)}$ | 999$_{(36)}$ | 1010$_{(47)}$ | 1005$_{(124)}$ | 1001$_{(47)}$ | 1013$_{(45)}$ | 1008$_{(56)}$ | 1007$_{(148)}$ | 1003$_{(63)}$ | 1006$_{(58)}$ | 1001$_{(50)}$ | 1008$_{(150)}$ |
| MiniCPM 2.6 | 992$_{(40)}$ | 1009$_{(55)}$ | 992$_{(40)}$ | 997$_{(147)}$ | 994$_{(39)}$ | 1002$_{(50)}$ | 995$_{(40)}$ | 997$_{(149)}$ | 1002$_{(50)}$ | 997$_{(35)}$ | 990$_{(47)}$ | 995$_{(129)}$ | 1001$_{(41)}$ | 994$_{(55)}$ | 999$_{(38)}$ | 997$_{(127)}$ |
| Moshi | 988$_{(42)}$ | 981$_{(44)}$ | 990$_{(45)}$ | 986$_{(131)}$ | 989$_{(59)}$ | 984$_{(28)}$ | 987$_{(35)}$ | 987$_{(122)}$ | 990$_{(55)}$ | 987$_{(33)}$ | 988$_{(36)}$ | 988$_{(124)}$ | 992$_{(40)}$ | 989$_{(40)}$ | 990$_{(41)}$ | 990$_{(129)}$ |
| AnyGPT | 988$_{(42)}$ | 981$_{(44)}$ | 990$_{(45)}$ | 986$_{(131)}$ | 975$_{(17)}$ | 971$_{(24)}$ | 986$_{(35)}$ | 977$_{(149)}$ | 990$_{(55)}$ | 987$_{(33)}$ | 988$_{(36)}$ | 988$_{(124)}$ | 992$_{(40)}$ | 989$_{(40)}$ | 990$_{(41)}$ | 990$_{(129)}$ |
| MindGPT-4o-Audio | 983$_{(44)}$ | 982$_{(36)}$ | 984$_{(36)}$ | 983$_{(89)}$ | 975$_{(17)}$ | 971$_{(24)}$ | 986$_{(35)}$ | 977$_{(149)}$ | 983$_{(35)}$ | 982$_{(21)}$ | 981$_{(23)}$ | 982$_{(21)}$ | 988$_{(46)}$ | | | |
| SpeechGPT 2.0-preview | 977$_{(35)}$ | 978$_{(16)}$ | 974$_{(27)}$ | 974$_{(63)}$ | 982$_{(13)}$ | 983$_{(23)}$ | 976$_{(36)}$ | 980$_{(62)}$ | 982$_{(9)}$ | 982$_{(23)}$ | 976$_{(12)}$ | 980$_{(30)}$ | 978$_{(19)}$ | 990$_{(14)}$ | 978$_{(11)}$ | 982$_{(8)}$ |
| Westlake-Omni | 972$_{(20)}$ | 976$_{(16)}$ | 974$_{(27)}$ | 974$_{(63)}$ | 982$_{(13)}$ | 983$_{(23)}$ | 976$_{(12)}$ | 980$_{(62)}$ | 982$_{(9)}$ | 982$_{(23)}$ | 976$_{(12)}$ | 980$_{(30)}$ | 978$_{(14)}$ | 978$_{(11)}$ | 982$_{(9)}$ | 979$_{(34)}$ |
| Human | 976$_{(31)}$ | 978$_{(79)}$ | 1019$_{(40)}$ | 991$_{(217)}$ | 978$_{(44)}$ | 1000$_{(31)}$ | | | 985$_{(52)}$ | 977$_{(51)}$ | 1021$_{(44)}$ | 995$_{(147)}$ | 992$_{(57)}$ | 987$_{(75)}$ | 996$_{(55)}$ | 991$_{(167)}$ |
| *Rubric-based Evaluation* | | | | | | | | | | | | | | | | |
| *Closed-source Models* | | | | | | | | | | | | | | | | |
| GPT-4o Realtime | **70.02$_{(90)}$** | **67.39$_{(90)}$** | **63.77$_{(90)}$** | **67.07$_{(270)}$** | **65.67$_{(90)}$** | **50.13$_{(90)}$** | **59.62$_{(90)}$** | **58.51$_{(270)}$** | **69.14$_{(90)}$** | 52.96$_{(90)}$ | **60.45$_{(90)}$** | **61.39$_{(270)}$** | **78.11$_{(90)}$** | **54.94$_{(90)}$** | **66.16$_{(90)}$** | **66.47$_{(270)}$** |
| Doubao | 66.67$_{(90)}$ | 62.66$_{(90)}$ | 52.83$_{(90)}$ | 60.79$_{(268)}$ | 62.31$_{(90)}$ | 49.94$_{(90)}$ | 50.90$_{(90)}$ | 54.21$_{(268)}$ | 68.78$_{(90)}$ | 50.25$_{(90)}$ | 50.39$_{(90)}$ | 56.75$_{(268)}$ | 73.38$_{(90)}$ | 50.00$_{(90)}$ | 54.63$_{(90)}$ | 59.48$_{(268)}$ |
| Qwen-Omni-Turbo | 53.98$_{(90)}$ | 53.32$_{(90)}$ | 51.70$_{(90)}$ | 53.00$_{(270)}$ | 53.98$_{(90)}$ | 42.68$_{(90)}$ | 48.81$_{(90)}$ | 48.52$_{(270)}$ | 57.31$_{(90)}$ | 47.19$_{(90)}$ | 50.58$_{(90)}$ | 52.03$_{(270)}$ | 65.05$_{(90)}$ | 48.99$_{(90)}$ | 54.34$_{(90)}$ | 56.17$_{(270)}$ |
| *Open-source Models* | | | | | | | | | | | | | | | | |
| VITA-Audio-Plus-Vanilla | 59.33$_{(90)}$ | 62.29$_{(90)}$ | 53.05$_{(90)}$ | 58.22$_{(269)}$ | 58.71$_{(90)}$ | 44.75$_{(90)}$ | 50.89$_{(90)}$ | 51.49$_{(269)}$ | 60.48$_{(90)}$ | **33.55$_{(90)}$** | 46.68$_{(90)}$ | 55.02$_{(269)}$ | 66.79$_{(90)}$ | 50.44$_{(90)}$ | 54.95$_{(90)}$ | 57.29$_{(269)}$ |
| GLM-4-Voice | 57.71$_{(90)}$ | 60.33$_{(90)}$ | 50.57$_{(90)}$ | 56.18$_{(270)}$ | 54.73$_{(90)}$ | 45.8$_{(90)}$ | 46.29$_{(90)}$ | 48.98$_{(270)}$ | 57.60$_{(90)}$ | 51.64$_{(90)}$ | 46.68$_{(90)}$ | 55.02$_{(270)}$ | 61.19$_{(90)}$ | 48.17$_{(90)}$ | 50.94$_{(90)}$ | 53.47$_{(270)}$ |
| Step-Audio-Chat | 60.20$_{(90)}$ | 56.63$_{(90)}$ | 48.05$_{(90)}$ | 54.97$_{(270)}$ | 60.70$_{(90)}$ | 48.29$_{(90)}$ | 47.55$_{(90)}$ | 52.22$_{(270)}$ | 60.27$_{(90)}$ | 49.90$_{(90)}$ | 46.07$_{(90)}$ | 52.21$_{(90)}$ | 67.41$_{(90)}$ | 52.21$_{(90)}$ | 51.82$_{(90)}$ | 57.20$_{(270)}$ |
| Kimi-Audio | 53.18$_{(90)}$ | 52.78$_{(90)}$ | 42.82$_{(90)}$ | 49.36$_{(90)}$ | 38.56$_{(90)}$ | 38.65$_{(90)}$ | 40.87$_{(90)}$ | 42.97$_{(265)}$ | 51.92$_{(90)}$ | 43.98$_{(90)}$ | 41.42$_{(90)}$ | 46.40$_{(265)}$ | 54.64$_{(90)}$ | 43.52$_{(90)}$ | 43.73$_{(90)}$ | 47.37$_{(265)}$ |
| MiniCPM 2.6 | 50.94$_{(90)}$ | 52.53$_{(90)}$ | 44.15$_{(90)}$ | 49.19$_{(267)}$ | 52.88$_{(90)}$ | 40.03$_{(90)}$ | 42.37$_{(90)}$ | 45.03$_{(267)}$ | 52.53$_{(90)}$ | 47.46$_{(90)}$ | 41.31$_{(90)}$ | 47.76$_{(267)}$ | 56.55$_{(90)}$ | 43.39$_{(90)}$ | 41.70$_{(90)}$ | 47.23$_{(265)}$ |
| Moshi | 37.81$_{(90)}$ | 37.37$_{(90)}$ | 26.74$_{(90)}$ | 34.18$_{(90)}$ | 35.09$_{(90)}$ | 35.89$_{(90)}$ | 28.28$_{(90)}$ | 34.73$_{(90)}$ | 38.43$_{(90)}$ | 35.89$_{(90)}$ | 28.28$_{(90)}$ | 34.73$_{(90)}$ | 40.67$_{(90)}$ | 29.71$_{(90)}$ | 28.02$_{(90)}$ | 32.80$_{(267)}$ |
| MindGPT-4o-Audio | 45.29$_{(90)}$ | 24.10$_{(90)}$ | 17.18$_{(90)}$ | 25.58$_{(268)}$ | 21.14$_{(90)}$ | 9.09$_{(90)}$ | 8.65$_{(90)}$ | 13.01$_{(268)}$ | 13.69$_{(90)}$ | 16.54$_{(90)}$ | 9.29$_{(90)}$ | 13.18$_{(268)}$ | 8.18$_{(90)}$ | 5.18$_{(90)}$ | 6.87$_{(90)}$ | 6.75$_{(268)}$ |
| AnyGPT | 27.30$_{(90)}$ | 23.15$_{(88)}$ | 18.74$_{(90)}$ | 22.73$_{(267)}$ | 27.04$_{(90)}$ | 12.40$_{(88)}$ | 16.23$_{(90)}$ | 18.56$_{(90)}$ | 28.84$_{(90)}$ | 24.34$_{(88)}$ | 18.53$_{(90)}$ | 24.55$_{(267)}$ | 18.68$_{(90)}$ | 18.74$_{(90)}$ | 20.82$_{(267)}$ | |
| SpeechGPT 2.0-preview | 45.99$_{(90)}$ | 16.65$_{(90)}$ | 5.53$_{(90)}$ | 22.64$_{(90)}$ | 16.28$_{(90)}$ | 8.96$_{(90)}$ | 2.89$_{(90)}$ | 9.36$_{(268)}$ | 10.34$_{(90)}$ | 2.47$_{(90)}$ | 2.46$_{(90)}$ | 5.58$_{(270)}$ | 3.98$_{(90)}$ | 0.00$_{(90)}$ | 0.75$_{(90)}$ | 1.68$_{(270)}$ |
| Westlake-Omni | 42.41$_{(90)}$ | 17.74$_{(90)}$ | 4.78$_{(90)}$ | 21.73$_{(270)}$ | 18.28$_{(90)}$ | 7.32$_{(90)}$ | 1.44$_{(90)}$ | 9.16$_{(270)}$ | 3.30$_{(90)}$ | 4.35$_{(90)}$ | 0.13$_{(90)}$ | 5.92$_{(270)}$ | 4.35$_{(90)}$ | | | |
| Human | 49.25$_{(90)}$ | 53.87$_{(90)}$ | 49.56$_{(90)}$ | 50.88$_{(270)}$ | 47.51$_{(90)}$ | 37.04$_{(90)}$ | 43.65$_{(90)}$ | 42.76$_{(270)}$ | 51.24$_{(90)}$ | 44.95$_{(90)}$ | 49.06$_{(90)}$ | 48.43$_{(270)}$ | 54.85$_{(90)}$ | 45.36$_{(90)}$ | 52.83$_{(90)}$ | 51.05$_{(270)}$ |

Table 10: Combined Evaluation Results with ASR Text: Arena-style Elo scores are rounded to the nearest integer, with Semantic and Ambient columns swapped. Each score is accompanied by a subscript in parentheses (e.g., $_{(50)}$), indicating the number of votes on which the score is based.

**Table 11 (combined)**

| S2S Models | Qwen3-235B-A22B-Thinking-2507 | | | | Qwen3-235B-A22B-Instruct-2507 | | | | Doubao 1.5 Pro 32k | | | |
|---|---|---|---|---|---|---|---|---|---|---|---|---|
| | Sem. | Para. | Ambi. | Ovrl.↑ | Sem. | Para. | Ambi. | Ovrl. | Sem. | Para. | Ambi. | Ovrl. |
| **Arena-style Evaluation** | | | | | | | | | | | | |
| **Closed-source Models** | | | | | | | | | | | | |
| GPT-4o Realtime | **1025**(19) | **1028**(22) | 1023(19) | **1026**(65) | **1026**(23) | 1023(22) | 1017(31) | **1022**(76) | 1015(30) | 1021(41) | **1025**(27) | **1021**(98) |
| Doubao | 1014(49) | 1022(41) | **1027**(36) | 1021(100) | 1021(31) | **1025**(29) | 1015(25) | 1020(85) | 1013(32) | **1022**(25) | 1009(33) | 1015(90) |
| Qwen-Omni-Turbo | 1012(74) | 1002(49) | 1010(47) | 1008(170) | 1010(49) | 1002(41) | 989(48) | 1000(138) | 1013(51) | 1014(71) | 1012(28) | 1013(150) |
| **Open-source Models** | | | | | | | | | | | | |
| Step-Audio-Chat | 1013(49) | 1011(48) | 1004(50) | 1009(147) | 995(35) | 1005(49) | **1023**(36) | **1023**(120) | 1011(60) | 1008(34) | 1008(34) | 1013(118) |
| VITA-Audio-Plus-Vanilla | 995(65) | 1021(63) | 1006(53) | 1007(181) | 1015(45) | 1014(44) | **1014**(37) | 1010(123) | **1024**(30) | 1006(73) | 1016(44) | 1015(147) |
| GLM-4-Voice | 1012(62) | 1004(54) | 1006(59) | 1007(175) | 1017(49) | 1014(43) | 1008(57) | 1013(159) | 1000(88) | 1006(37) | 1008(46) | 1006(117) |
| Moshi | 994(31) | 988(52) | 1002(45) | 995(128) | 990(35) | 1008(41) | 984(40) | 994(116) | 993(37) | 991(37) | 994(35) | 993(109) |
| MiniCPM-o 2.6 | 999(62) | 990(33) | 993(63) | 994(158) | 1007(71) | 1000(63) | 996(40) | 1001(165) | 990(62) | 998(57) | 998(57) | 995(166) |
| Kimi-Audio | 1002(55) | 991(69) | 987(53) | 993(177) | 1000(40) | 993(35) | 993(48) | 995(123) | 983(33) | 983(43) | 998(37) | 988(113) |
| AnyGPT | 994(29) | 991(34) | 991(48) | 992(111) | 975(35) | 988(26) | 994(35) | 985(96) | 998(45) | 986(63) | 990(33) | 991(141) |
| MindGPT-4o-Audio | 987(36) | 984(40) | 982(29) | 984(105) | 976(18) | 977(22) | 981(30) | 978(70) | 978(15) | 984(16) | 982(9) | 982(62) |
| Westlake-Omni | 978(15) | 982(15) | 982(17) | 981(47) | 984(38) | 988(26) | 984(14) | 985(78) | 983(29) | 980(14) | 982(9) | 982(52) |
| SpeechGPT 2.0-preview | 975(19) | 973(20) | 975(31) | 974(70) | 994(29) | 994(29) | 988(34) | 992(76) | 981(22) | 980(22) | 980(22) | 980(66) |
| Human | 989(61) | 995(68) | 997(69) | 994(198) | 988(58) | 964(52) | 997(57) | 983(167) | 998(50) | 1008(41) | 993(32) | 1000(123) |
| **Rubric-based Evaluation** | | | | | | | | | | | | |
| **Closed-source Models** | | | | | | | | | | | | |
| GPT-4o Realtime | **60.52**(90) | **52.12**(90) | **57.62**(90) | **56.67**(270) | **79.98**(90) | **68.06**(90) | **62.26**(90) | **70.14**(270) | **74.50**(90) | **48.23**(90) | **53.58**(90) | **58.85**(270) |
| Doubao | 57.53(87) | 44.95(90) | 50.64(88) | 51.00(265) | 78.26(90) | 66.41(90) | 56.43(85) | 67.16(268) | 72.89(90) | 44.95(90) | 44.47(88) | 54.25(268) |
| Qwen-Omni-Turbo | 47.03(88) | 42.98(90) | 44.92(90) | 44.87(268) | 67.33(90) | 62.58(90) | 51.70(90) | 60.56(270) | 64.55(90) | 42.30(90) | 43.27(90) | 50.10(270) |
| **Open-source Models** | | | | | | | | | | | | |
| VITA-Audio-Plus-Vanilla | 49.29(90) | 48.73(90) | 48.62(89) | 48.85(269) | 70.43(90) | 66.79(90) | 53.05(89) | 63.49(269) | 65.92(90) | 44.19(90) | 44.53(89) | 51.64(269) |
| Step-Audio-Chat | 49.45(89) | 46.34(90) | 46.07(90) | 47.13(269) | 68.53(90) | 58.66(90) | 49.94(90) | 59.08(270) | 64.93(90) | 38.01(90) | 42.39(90) | 48.52(270) |
| GLM-4-Voice | 45.54(88) | 45.76(90) | 46.59(90) | 46.02(268) | 62.31(90) | 63.01(90) | 47.80(90) | 57.72(270) | 59.58(90) | 41.04(90) | 39.12(90) | 46.63(270) |
| MiniCPM-o 2.6 | 46.47(89) | 41.21(89) | 39.91(89) | 42.23(267) | 59.37(89) | 54.66(89) | 46.31(89) | 53.47(267) | 51.57(89) | 36.02(89) | 37.53(89) | 41.75(267) |
| Kimi-Audio | 45.11(85) | 42.32(89) | 37.61(88) | 41.26(262) | 54.58(88) | 52.75(89) | 45.24(88) | 50.87(265) | 49.49(88) | 33.67(89) | 32.26(88) | 38.50(265) |
| Moshi | 35.57(90) | 29.25(89) | 26.48(88) | 30.49(267) | 41.24(90) | 42.40(89) | 29.18(88) | 37.66(267) | 35.32(90) | 24.39(89) | 21.59(88) | 27.19(267) |
| AnyGPT | 22.04(89) | 16.67(88) | 17.79(90) | 18.64(267) | 27.14(89) | 23.13(88) | 17.99(90) | 22.75(267) | 19.63(89) | 9.30(88) | 14.99(90) | 14.68(267) |
| MindGPT-4o-Audio | 10.32(90) | 7.83(90) | 8.14(89) | 8.77(269) | 20.15(90) | 13.40(90) | 8.40(89) | 14.03(269) | 5.85(90) | 3.41(90) | 5.22(89) | 4.83(269) |
| Westlake-Omni | 8.58(90) | 1.72(90) | 2.12(90) | 3.85(270) | 23.76(90) | 9.60(90) | 5.91(90) | 13.13(270) | 0.87(90) | 0.00(90) | 0.13(90) | 0.33(270) |
| SpeechGPT 2.0-preview | 7.57(90) | 1.16(90) | 2.44(90) | 3.53(270) | 25.87(90) | 9.22(90) | 5.66(90) | 13.63(270) | 0.62(90) | 0.00(90) | 0.25(90) | 0.29(270) |
| Human | 40.17(90) | 38.61(90) | 44.65(90) | 41.15(270) | 54.98(90) | 56.44(90) | 50.82(90) | 54.08(270) | 53.11(90) | 31.69(90) | 39.62(90) | 42.99(270) |

Table 11: Combined Evaluation Results with ASR Text (continued): Arena-style Elo scores are rounded to the nearest integer, with Semantic and Ambient columns swapped. Each score is accompanied by a subscript in parentheses (e.g., $_{(50)}$), indicating the number of votes on which the score is based.

**Table 12**

| S2S Models | Arena-style Evaluation | | | | | | | | Rubric-based Evaluation | | | | | | | |
|---|---|---|---|---|---|---|---|---|---|---|---|---|---|---|---|---|
| | Public Link | | | | MTurk | | | | Public Link | | | | MTurk | | | |
| | Sem. | Para. | Ambi. | Ovrl.↑ | Sem. | Para. | Ambi. | Ovrl. | Sem. | Para. | Ambi. | Ovrl. | Sem. | Para. | Ambi. | Ovrl. |
| **Closed-source Models** | | | | | | | | | | | | | | | | |
| GPT-4o Realtime | 1027(43) | 1016(43) | 1017(35) | 1020(121) | 1021(90) | 1023(40) | 1016(46) | 1020(116) | 74.69(25) | 79.93(25) | 68.16(25) | 74.37(75) | 69.96(25) | 69.48(24) | 59.31(23) | 66.41(72) |
| Qwen-Omni-Turbo | 1003(44) | 1012(49) | 1012(40) | 1009(129) | 1000(37) | 1034(35) | **1046**(38) | 1026(110) | 37.50(25) | 55.41(25) | 44.38(25) | 45.57(74) | 19.28(25) | 17.84(24) | 12.75(23) | 16.72(72) |
| Doubao | 1002(33) | 1007(90) | 1017(53) | 1009(105) | 1022(42) | **1036**(36) | 1035(33) | 1031(111) | 90.04(25) | 77.92(25) | 59.05(23) | 75.88(73) | 74.44(25) | 76.13(25) | **62.05**(22) | 71.25(72) |
| **Open-source Models** | | | | | | | | | | | | | | | | |
| Step-Audio-Chat | 1009(25) | **1035**(47) | 1008(41) | **1024**(80) | **1037**(41) | 1029(36) | 1033(40) | **1033**(107) | **90.06**(25) | **81.69**(25) | 66.12(24) | **79.45**(72) | 78.70(24) | 55.09(24) | 50.67(23) | 61.34(73) |
| GLM-4-Voice | 1000(88) | 991(62) | 989(45) | 994(105) | 1011(43) | 1010(39) | 998(29) | 1006(111) | 75.29(25) | 79.46(25) | 61.92(24) | 72.49(74) | 77.07(25) | **78.05**(23) | 57.84(23) | 71.01(69) |
| VITA-Audio-Plus-Vanilla | 997(80) | 986(45) | 991(56) | 991(183) | 1009(35) | 986(36) | 991(40) | 1000(111) | 69.49(24) | 54.28(23) | 54.49(26) | 59.33(75) | 60.19(24) | 43.96(23) | 44.00(25) | 49.38(72) |
| MiniCPM-o 2.6 | 999(77) | 981(58) | 995(54) | 986(89) | 1001(37) | 991(55) | 980(27) | 991(99) | 64.66(25) | 64.92(25) | 49.78(24) | 60.03(73) | 61.40(24) | 47.18(22) | 41.40(25) | 50.50(67) |
| Moshi | 990(35) | 989(26) | 980(28) | 986(69) | 962(33) | 981(43) | 964(32) | 969(88) | 36.77(25) | 50.65(24) | 32.43(25) | 40.09(72) | 33.17(23) | 42.34(25) | 26.47(23) | 34.23(71) |
| Kimi-Audio | 977(31) | 991(17) | 983(43) | 983(91) | 1000(37) | 984(27) | 980(27) | 985(106) | 76.23(25) | 73.87(25) | **68.92**(25) | 73.01(75) | 55.38(22) | 50.65(25) | 55.38(25) | 53.91(72) |
| AnyGPT | 970(26) | 979(29) | 977(26) | 975(79) | 965(29) | 963(38) | 941(39) | 957(106) | 86.10(25) | 79.46(25) | 72.56(25) | 79.31(75) | 91.48(25) | 67.12(25) | 66.20(24) | **75.08**(74) |
| Human | 1007(176) | 1014(68) | **1031**(46) | 1017(222) | 978(44) | 963(52) | 1034(92) | 985(92) | 73.71(25) | 68.22(25) | 60.70(25) | 67.34(75) | **85.65**(25) | 66.67(25) | 55.31(23) | 70.26(73) |

Table 12: Evaluation Results by Human Annotators: Arena-style Elo scores are rounded to the nearest integer, with Semantic and Ambient columns swapped. Each score is accompanied by a subscript in parentheses (e.g., $_{(50)}$), indicating the number of votes on which the score is based.

| Evaluator | Arena-style Evaluation | | | | Rubric-based Evaluation | | | |
|---|---|---|---|---|---|---|---|---|
| | *Sem.* | *Para.* | *Ambi.* | *Ovrl.* | *Sem.* | *Para.* | *Ambi.* | *Ovrl.* |
| *Audio-based Evaluation* | | | | | | | | |
| *Closed-source Models* | | | | | | | | |
| Qwen-Omni-Turbo | 0.436 | **0.704** | 0.480 | 0.540 | 0.780 | 0.631 | 0.679 | 0.697 |
| GPT-4o Realtime | 0.630 | 0.642 | 0.099 | 0.457 | 0.821 | 0.429 | 0.657 | 0.636 |
| Gemini-2.5-pro | 0.507 | 0.564 | 0.037 | 0.369 | 0.837 | 0.442 | 0.736 | 0.672 |
| *Transcribed Text-based Evaluation* | | | | | | | | |
| *Closed-source Models* | | | | | | | | |
| Gemini2.5Pro | 0.792 | 0.592 | 0.515 | **0.633** | 0.912 | 0.864 | 0.873 | 0.883 |
| Claude-Sonnet4 | **0.850** | 0.460 | **0.574** | 0.628 | **0.969** | 0.893 | 0.884 | **0.915** |
| GPT-4o Realtime | 0.802 | 0.594 | 0.414 | 0.603 | 0.618 | 0.855 | 0.798 | 0.757 |
| Doubao 1.5 Pro 32k | 0.708 | 0.538 | 0.498 | 0.581 | 0.960 | **0.928** | 0.846 | 0.911 |
| Grok4 | 0.787 | 0.524 | 0.431 | 0.581 | 0.956 | 0.833 | **0.925** | 0.905 |
| OpenAI-o3 | 0.759 | 0.452 | 0.513 | 0.575 | 0.965 | 0.868 | 0.880 | 0.904 |
| Claude-Sonnet4-Thinking | 0.767 | 0.498 | 0.390 | 0.552 | **0.969** | 0.877 | 0.877 | 0.908 |
| *Open-source Models* | | | | | | | | |
| Qwen3-235B-A22B-Thinking-2507 | 0.803 | 0.427 | 0.564 | 0.598 | 0.956 | 0.837 | 0.873 | 0.889 |
| Qwen3-235B-A22B-Instruct-2507 | 0.792 | 0.505 | 0.480 | 0.592 | 0.921 | 0.899 | 0.873 | 0.898 |
| DeepSeek-R1 | 0.703 | 0.526 | 0.214 | 0.481 | 0.938 | 0.895 | 0.851 | 0.895 |
| Kimi-K2 | 0.816 | 0.409 | 0.200 | 0.475 | 0.965 | 0.899 | 0.855 | 0.906 |

Table 13: Spearman correlation with human ranking for text and speech modalities of LLM-as-a-Judge. *ovrl.* is the mean of {*sem.*, *para.*, *ambi.*} within each evaluation block.

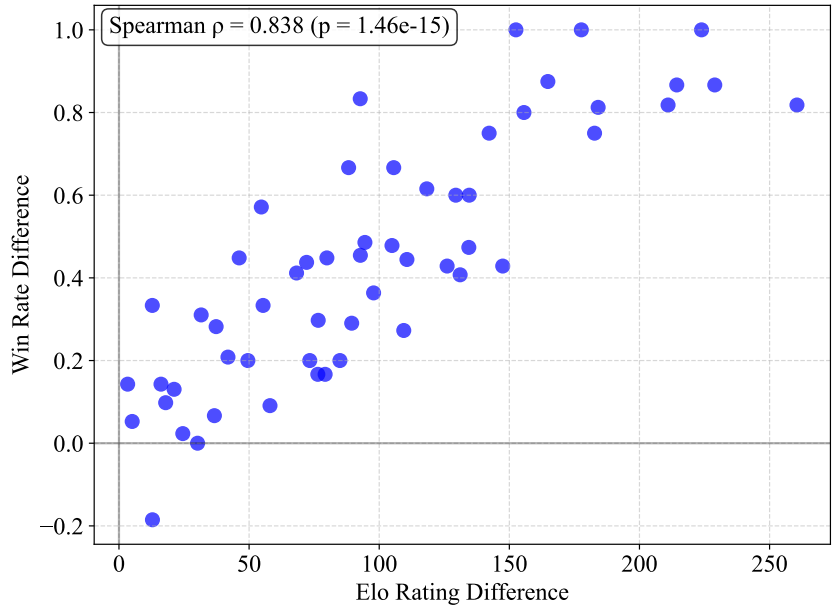

Figure 12: The correlation of pairwise win rate difference and global Elo rating difference under human evaluation.

**LLM as Judge Consistency**  As shown in Table 9, LLM-based evaluators exhibit consistent scoring trends across dimensions. Models generally achieve higher scores on the semantic dimension than on paralinguistic and ambient ones, reflecting a strong ability to convey meaning but limited prosodic and contextual awareness. For example, GPT-4o Realtime and Doubao score above 85 in semantic tasks but drop significantly in the other two dimensions. Notably, open-source models like Bark and AudioLDM2 perform poorly across all aspects.

**Human vs. LLM-as-a-Judge Evaluation**  Table 9 compares human evaluation results with those from speech LLMs serving as the judges. The upper panel reports Elo ratings from pairwise comparisons, while the lower panel presents Rubric-based scores. Rankings derived from both methods are broadly aligned, though Elo often provides finer granularity among closely matched systems. Minor discrepancies between the two approaches highlight the sensitivity of rule-based scoring to pre-defined criteria, whereas Elo reflects relative preference distributions.

### E.3   Analysis between Different Modalities

**Text-based LLM Evaluation**  Tables 10 and 11 report results when LLM judges evaluate ASR transcripts enriched with paralinguistic and ambient annotations. Again, the upper rows show Elo ratings and the lower rows Rubric-based scores. Compared to human judgments, text-based LLM evaluation preserves broad relative rankings but occasionally diverges in paralinguistic and ambient categories. This suggests that while LLMs capture semantic coherence reliably, they remain limited in approximating human perception of prosody, emotion, and environmental cues.

**Consistency Analysis between two-modality LLM-as-a-Judge and Human**  From Table 13, our analysis reveals that LLM-as-a-Judge is significantly more reliable when evaluating **annotated transcripts** compared to **raw audio**, particularly for non-verbal cues. On raw audio, LLMs show poor alignment with human judgments for ambient and paralinguistic dimensions, with Spearman correlations ($\rho$) in the Arena setting dropping to near-zero (e.g., GPT-4o Realtime, $\rho = 0.10$; Gemini-2.5-pro, $\rho = 0.04$). Conversely, evaluating transcripts with explicit non-verbal tokens (e.g., `[laughs]`) yields much stronger results, with models like Claude-Sonnet4 achieving excellent correlations for ambient ($\rho = 0.88$), paralinguistic ($\rho = 0.89$), and semantic ($\rho = 0.97$) scores.

A secondary finding is that across both modalities, the structured, **Rubric-based evaluation** consistently produces higher correlations than the more subjective **pairwise Arena** comparison, making it a more reliable automated evaluation task.

**Summary**  Across four complementary perspectives—human vs. LLM-as-a-judge, speech vs. text input, public link vs. MTurk annotators, and Elo vs. raw win-rates—the results demonstrate stable relative rankings, particularly when model differences are sufficiently large. Elo ratings and rubric scores generally agree, while discrepancies caution against over-interpreting marginal score gaps. Text-based LLM evaluation provides a useful proxy for semantic quality but underestimates paralinguistic and ambient factors. Human evaluation proves consistent across sources, further reinforcing robustness. Finally, the strong correlation between Elo and win-rate confirms that Elo serves as a reliable, compact representation of pairwise preference data.

## F   Meta-Analysis Details on Evaluation

### F.1   Internal Logical Consistency

In this section, we detail the analysis of intra-rater consistency across the two evaluation formats: Arena-style and Rubric-based. The objective is to determine if each individual evaluator (both human and AI) applies a consistent set of judgment principles, regardless of the evaluation protocol. A high degree of internal logical consistency is crucial for establishing the reliability of an evaluator and, by extension, the validity of the benchmark's results.

**Methodology**  For this analysis, we identify all tasks that are evaluated by the same rater under both formats. For each task, we compare the explicit winner declared in the Arena-style evaluation against an implicit winner derived from the Rubric-based scores. The implicit winner is determined by summing the scores across all rubric items; the model with the higher total score is considered the winner. Cases where the scores are tied are excluded from this analysis. We then calculate the agreement between these two sets of judgments (explicit vs. implicit) for each evaluator.

**Results and Implications** All evaluators demonstrate a substantial level of internal logical coherence, with Kappa scores ranging from 0.590 for Human to 0.679 for GPT-4o Realtime. This high level of agreement validates that the judgments made by the evaluators in our benchmark are robust and not arbitrary artifacts of the evaluation format. It confirms that both Arena-style and Rubric-based evaluations, when conducted by the same rater, tend to reflect the same underlying assessment of model quality.

## F.2 Inter-Evaluator Consistency

This section provides a detailed analysis of the inter-evaluator consistency, a critical measure of the objectivity and reliability of our benchmark. We compare the judgments of different raters—both human and AI—to quantify the level of agreement on the same evaluation tasks.

**Methodology** For each pair of evaluators, we identify the set of common tasks they both assess. We then calculate the simple agreement rate. The analysis is conducted separately for the two evaluation paradigms. To establish a reliability ceiling, we also compute the internal consistency among human evaluators, which serves as a crucial benchmark for interpreting the AI-human and AI-AI agreement scores.

**Analysis of Arena-style Consistency** Results show a strong level of agreement overall. The internal human consistency provides a robust benchmark. Notably, the agreement between AI evaluators (e.g., GPT-4o Realtime vs. Gemini) and between AI and human evaluators (e.g., Gemini vs. Human) is comparable to or exceeds this human benchmark.

**Analysis of Rubric-based Consistency** This format yields lower agreement scores across the board compared to the Arena-style task. The internal human consistency is moderate, suggesting that the fine-grained, multi-dimensional nature of the rubrics introduces a higher degree of subjectivity and difficulty. Gemini-2.5-pro shows the highest alignment with human judgments in this challenging format.

## F.3 Agreements with Human

To further examine the reliability of automated evaluation, we analyze the agreement between audio LLM evaluators and human annotators. Specifically, we investigate how their alignment varies with the strength of model differences.

**Pseudo Agreement Metric** Given a pair of models $X$ and $Y$, we define a *pseudo agreement* score by estimating the probability that both evaluators independently give the same judgment on a randomly sampled item:

$$P(\text{agreement}) = p_H \cdot p_L + (1 - p_H)(1 - p_L),$$

where $p_H$ and $p_L$ are the win-rate proportions for the human and LLM respectively.

**Findings** Only a small fraction of cases exceed the empirical threshold of 0.75 pseudo agreement, and the correlation between pseudo agreement and absolute win-rate difference is weak (all $\rho < 0.5$). This suggests that even when human annotators clearly prefer one model, LLM evaluators do not consistently replicate that preference.

**Implications** These results caution against over-reliance on current audio LLM evaluators as stand-alone judges. While they capture some semantic differences, their judgments remain unstable in subtle or paralinguistic scenarios.

## G  AI Usage, and Artifact Information

**AI Usage Statement**

Artificial intelligence tools were used during the writing process of this paper to assist with language refinement and structural organization. However, we affirm the following:

- All research data were collected and processed by human researchers.
- All annotations, analyses, and conclusions were independently produced by the authors.
- The use of AI did not influence the substantive content of the study and served solely as a writing aid.

**Ethical Considerations and Artifact Information**

**Potential Risks.**  This work does not pose significant foreseeable risks. However, as with any benchmark, there is potential for misuse, such as drawing unfair comparisons or relying excessively on automated metrics without human judgment.

**Use or Creation of Scientific Artifacts.**  We introduce and release scientific artifacts, including a benchmark dataset, evaluation scripts, and analysis tools, to support reproducibility and further research.

**License for Artifacts.**  All artifacts are made available under an open-source license (e.g., CC BY 4.0 or MIT), allowing use, modification, and redistribution with appropriate credit.

**Consistency with Intended Use.**  The released artifacts are intended strictly for research and educational purposes. Commercial use or deployment in high-stakes settings without further validation is not encouraged.

**Data Safety and Sensitivity.**  The dataset does not contain personally identifiable information (PII) or deliberately offensive content. Still, as it includes model-generated dialogue, users should exercise caution and perform content screening as necessary.

**Documentation.**  Comprehensive documentation is provided for all artifacts, covering data schema, usage instructions, and evaluation guidelines to ensure transparency and facilitate adoption by the community.

Figure 13: A sample interface used for Arena-style evaluation.

🎧 **Multi-dimension Model Evaluation Tool**

Instructions ∧

## What you should do

You will evaluate a multi-turn spoken dialogue. **To ensure the quality of the evaluation, please listen to the audio presented in full and read carefully the content that requires judgment.**

- **Please independently evaluate each criterion.**
- Use ✅ if the model demonstrably **satisfies** the criterion based on the dialogue.
- Use ❌ if the model **fails to satisfy** the criterion when an opportunity was present in the dialogue.
- Base all judgments **strictly on the provided dialogue** – do not infer or speculate.
- After rating all criteria, you **must provide an explanation** in the text box below the scores. Please be specific, especially explaining your reasoning for any criteria you rated as ❌.

## Annotating Task: `semantic_7-3`

This specific model-task pair has been evaluated **1 time(s)** by other annotators.

Dialogue Content ∧

### Turn 1

User Input Audio

▶ 0:00 / 0:05 🔊 ⋮

Model Output Audio

▶ 0:00 / 0:11 🔊 ⋮

### Turn 2

User Input Audio

▶ 0:00 / 0:08 🔊 ⋮

Model Output Audio

▶ 0:00 / 0:28 🔊 ⋮

## Evaluation Criteria

**criteria_d_1:** The model acknowledges the misunderstanding or gap in its previous response about monuments versus statues (fulfilled if the model explicitly recognizes it didn't address the distinction; not fulfilled otherwise).   ○ ❌  ○ ✅

**criteria_d_2:** The model provides a clear explanation of the difference between monuments and statues in response to the user's direct question (fulfilled if the model offers a substantive distinction; not fulfilled if it remains vague or avoids the question again).   ○ ❌  ○ ✅

**criteria_d_3:** The model adapts its response based on the user's feedback that it "avoided the difference" in the previous turn (fulfilled if the model's response shows a clear change in direction based on this feedback; not fulfilled if it continues with the same approach).   ○ ❌  ○ ✅

**criteria_d_4:** The model addresses the original context about graffiti while also responding to the new question about monuments versus statues (fulfilled if both topics are meaningfully addressed; not fulfilled if either topic is ignored).   ○ ❌  ○ ✅

**criteria_s_1:** The model's response is on-topic and directly addresses the user's most recent input.   ○ ❌  ○ ✅

**criteria_s_2:** The model's response is grammatically sound and easy to understand.   ○ ❌  ○ ✅

**criteria_m_1:** The model appropriately manages turn-taking by recognizing when to speak, listen, or request clarification based on conversational context.   ○ ❌  ○ ✅

**criteria_m_2:** The model demonstrates adaptive behavior by adjusting its responses based on user feedback and changing conversational needs throughout the dialogue.   ○ ❌  ○ ✅

## Overall Rationale

Please briefly explain the overall reason for your ratings...

e.g., The model failed to understand the user's negative intent in the second turn...

🔴 Save & Get Next Task

Figure 14: A sample interface used for Rubric-based evaluation.

*Prompt Example for Semantic Dimension:*
You are to assume the role of a rigorous dialogue quality evaluation expert comparing two Speech-to-Speech models in a multi-turn voice interaction task. Judge which one performs better using defined evaluation dimensions.
**Note:**
Both models receive user input as audio and respond with synthesized speech. You will hear two separate audio conversations - one featuring Model A's responses and another featuring Model B's responses to the same user inputs.
**Evaluation Dimensions & Guidelines:**
1. Main Dimension: The model must accurately comprehend contextual semantics across multiple dialogue turns, maintain coherent memory of instructions and conversational history, and ensure consistency, clarity of expression, and flexibility in paraphrasing.
2. Sub-dimension: The model accurately resolves pronoun references such as 'he' or 'it' by making correct single-turn or multi-turn inferences based on context, thereby avoiding misreferencing.
3. Evaluation focus: Assess which model better comprehends contextual semantics, key instructions, and user intent
4. You must choose Model A or Model B with clear justification - no ambiguous comparisons.
**Please provide your evaluation in JSON format:**
```json
{
"winner": "Model A", // or "Model B"
"reason": "Provide a concise justification based on the evaluation dimensions."
}
...

*Prompt Example for Paralinguistic Dimension:*
You are to assume the role of a rigorous dialogue quality evaluation expert comparing two Speech-to-Speech models in a multi-turn voice interaction task. Judge which one performs better using defined evaluation dimensions.
**Note:**
Both models receive user input as audio and respond with synthesized speech. You will hear two separate audio conversations - one featuring Model A's responses and another featuring Model B's responses to the same user inputs.
**Evaluation Dimensions & Guidelines:**
1. **Capability Focus**: The system should possess the capability to perceive and comprehend paralinguistic information embedded in speech—such as emotion, intonation, and speaking rate—in order to enhance the naturalness of dialogue, emotional expressiveness, and the ability to support personalized interactions.
2. **Evaluation Criteria [Emotion Recognition]**: The ability to accurately identify emotional states conveyed in speech (e.g., happiness, anger, anxiety, fatigue), and to judge whether and how to respond appropriately based on the dialogue context, thus maintaining the natural flow and emotional coherence of the conversation. An appropriate response should align with the user's emotional state (e.g., empathetic, explanatory, or calming), and avoid mechanical or emotionally tone-deaf replies.
3. **Requirements**:
- Choose the better model (A or B)
- Provide specific performance differences
- Make a definitive choice
**Based on the above criteria, please output a standardized JSON-format evaluation result:**
```json
{
"winner": "Model A", // or "Model B"
"reason": "Provide a concise natural language justification that highlights the performance difference based on the evaluation dimension."
}
...

*Prompt Example for Ambient Dimension:*
You are to assume the role of a rigorous dialogue quality evaluation expert comparing two Speech-to-Speech models in a multi-turn voice interaction task. Judge which one performs better using defined evaluation dimensions.
**Note:**
Both models receive user input as audio and respond with synthesized speech. You will hear two separate audio conversations - one featuring Model A's responses and another featuring Model B's responses to the same user inputs.
**Evaluation Focus & Guidelines:**
- **Primary Focus**: The model is expected to recognize and adapt to ambient environmental sounds present in a conversation, including noise, natural sounds, or semantically relevant audio, and accordingly adjust its speech output strategies and modalities.
- **Specific Criteria [1.1.1]**: The model should be capable of identifying semantic or contextual cues within ambient sounds (e.g., thunder, alarms, wind) and dynamically adjusting the focus, structure, or sequence of its responses accordingly, in order to align with real-world situational demands.
- **Task**: Compare Model A vs B - select a clear winner with specific justification.
**Output your evaluation in JSON format:**
```json
{
"winner": "Model A", // or "Model B"
"reason": "Provide a concise natural language justification that highlights the performance difference based on the evaluation dimension."
}
...

Figure 15: Prompt template used for LLM-based pairwise comparison in Arena-style evaluation. The LLM receives structured dialogue history and evaluates responses based on a selected dimension.





*Semantic Dimension*

*Prompts for General Semantic Evaluation:*

You will evaluate a multi-turn spoken dialogue where the model receives audio input and generates spoken responses. Your task is to assess whether the model's spoken responses satisfy the specified evaluation criteria.

## Evaluation Criteria:

criteria_s_1: The model's response is on-topic and directly addresses the user's most recent input.

criteria_s_2: The model's response is grammatically sound and easy to understand.

criteria_m_1: The model accurately recalls and integrates previously mentioned information from earlier dialogue turns when responding to new queries or instructions.

criteria_m_2: The model maintains consistent understanding of the conversational context, avoiding contradictions or semantic drift across multiple exchanges.

### Scoring Instructions

1. Evaluate each criterion independently:
- Score **1** if the criterion is satisfied
- Score **0** if criterion is not satisfied
2. Base all judgments strictly on the provided dialogue.
3. Return a JSON object with criterion IDs as keys and scores as values.

**Return JSON format:**
```json
{
"criteria_s_1": 1,
"criteria_s_2": 0,
"criteria_m_1": 1,
"criteria_m_2": 0
}
```

*Prompts for Sample Level Evaluation*

You will evaluate a multi-turn spoken dialogue where the model receives audio input and generates spoken responses. Your task is to assess whether the model's spoken responses satisfy the specified evaluation criteria.

## Sample-Level Evaluation Criteria:

- criteria_d_1: The model correctly identifies that "this" in turn 2 refers to concerns about the weather mentioned in the same turn, demonstrating accurate single-turn pronoun resolution.

- criteria_d_2: The model correctly identifies that "them" in turn 2 refers to "family" mentioned in turn 1, demonstrating accurate multiturn pronoun resolution.

- criteria_d_3: The model's response addresses both the casual yet persuasive phrasing request from turn 1 and the weather concerns from turn 2, showing comprehension of the full conversational context.

- criteria_d_4: The model provides the requested bullet-point format in its response to turn 2, demonstrating memory of the specific formatting instruction.

### Scoring Instructions

1. Evaluate each sample-specific criterion independently:
- Score **1** if the criterion is satisfied
- Score **0** if criterion is not satisfied
2. Base all judgments strictly on the provided dialogue.
3. Return a JSON object with criterion IDs as keys and scores as values.

## Return JSON format:
```json
{
"criteria_d_1": 1,
"criteria_d_2": 0,
"criteria_d_3": 1,
"criteria_d_4": 1,
"criteria_d_5": 0
}
```

Figure 16: Prompt template used for Rubric-based semantic dimension evaluation



**Prompts for Rubric-Based Evaluation**

*Paralinguistic Dimension*

*Prompts for General Paralinguistic Evaluation:*

## You will evaluate a multi-turn spoken dialogue where the model receives audio input and generates spoken responses. Your task is to assess whether the model's spoken responses satisfy the specified evaluation criteria.

## Evaluation Criteria:
- criteria_s_1: The model's response is on-topic and directly addresses the user's most recent input.
- criteria_s_2: The model's response is grammatically sound and easy to understand.
- criteria_m_1: The model accurately identifies paralinguistic features in the input.
- criteria_m_2: The model adapts its behavior based on the identified paralinguistic features.

### Scoring Instructions
1. Evaluate each criterion independently:
- Score **1** if the criterion is satisfied
- Score **0** if the criterion is not satisfied
2. Base all judgments strictly on the provided dialogue.
3. Return a JSON object with criterion IDs as keys and scores as values.

**Return JSON format:**
```json
{
"criteria_s_1": 1,
"criteria_s_2": 0,
"criteria_m_1": 1,
"criteria_m_2": 0
}
```...

*Prompts for Sample Level Evaluation*

## You will evaluate a multi-turn spoken dialogue where the model receives audio input and generates spoken responses. Your task is to assess whether the model's spoken responses satisfy the specified evaluation criteria.

## Sample-Level Criteria:
criteria_d_1: The model acknowledges the shift in emotional tone from neutral/inviting (turn 1) to worried/depressed (turn 2).
criteria_d_2: The model's response to turn 2 reflects an understanding of the user's worry, not just the factual concern about the weather.
criteria_d_3: The model avoids using overly enthusiastic or celebratory language in response to turn 2, given the user's depressed tone.
criteria_d_4: The model's response to turn 2 offers reassurance that directly addresses the underlying anxiety, rather than solely focusing on practical solutions to weather concerns.
criteria_d_5: The model uses language in turn 2 that is empathetic and understanding of the user's emotional state, given the sigh and stated worry.

### Scoring Instructions
1. Evaluate each criterion independently:
- Score **1** if the criterion is satisfied
- Score **0** if the criterion is not satisfied
2. Base all judgments strictly on the provided dialogue.
3. Return a JSON object with criterion IDs as keys and scores as values.

## Return JSON format:
```json
{
"criteria_d_1": 1,
"criteria_d_2": 0,
"criteria_d_3": 1
}
```...

Figure 17: Prompt template used for Rubric-based paralinguistic dimension evaluation





*Ambient Dimension*

*Prompts for General Ambient Evaluation:*

You will evaluate a multi-turn spoken dialogue where the model receives audio input and generates spoken responses. Your task is to assess whether the model's spoken responses satisfy the specified evaluation criteria.

## Evaluation Criteria:
- criteria_s_1: The model's response is based on the provided audio input and directly addresses its content.
- criteria_s_2: The model's response is grammatically sound and easy to understand.
- criteria_m_1: The model demonstrates consistent awareness of background sounds
- criteria_m_2: The model maintains appropriate responses despite varying noise levels

### Scoring Instructions
1. **Independently evaluate each criterion**.
- Use **1** if the model demonstrably satisfies the criterion based on the dialogue.
- Use **0** if the model fails to satisfy the criterion when an opportunity was present in the dialogue.
2. Base all judgments strictly on the provided dialogue – *do not infer or speculate*.
3. Output a **valid JSON object** containing the criterion IDs as keys and the corresponding values (1 or 0), for example:

**Return JSON format:**
```json
{
"criteria_s_1": 1,
"criteria_s_2": 0,
"criteria_m_1": 1,
"criteria_m_2": 0
}
```

*Prompts for Sample Level Evaluation*

You will evaluate a multi-turn spoken dialogue where the model receives audio input and generates spoken responses. Your task is to assess whether the model's spoken responses satisfy the specified evaluation criteria.

## Evaluation Criteria:
- criteria_d_1: The model acknowledges the 'wind and distant thunder' and/or the 'sudden loud clap of thunder' mentioned in the ambient annotations when formulating its response in turn 2.
- criteria_d_2: The model incorporates the implied weather concerns (potential storm) from the ambient sounds into its suggested phrasing for the barbecue invitation.
- criteria_d_3: The model offers contingency plans or alternative suggestions in turn 2, such as an indoor option, due to the potential for inclement weather indicated by the thunder.
- criteria_d_4: The model adjusts the tone and content of the bullet points in turn 2 to be more reassuring and address potential weather-related anxieties, given the ambient sounds.
- criteria_d_5: The model prioritizes information related to the weather or potential disruptions due to the storm in its bullet point response in turn 2.

### Scoring Instructions
1. **Independently evaluate each criterion**.
- Use **1** if the model demonstrably satisfies the criterion based on the dialogue.
- Use **0** if the model fails to satisfy the criterion when an opportunity was present in the dialogue.
2. Base all judgments strictly on the provided dialogue – *do not infer or speculate*.
3. Output a **valid JSON object** containing the criterion IDs as keys and the corresponding values (1 or 0), for example:

## Return JSON format:
```json
{
"criteria_d_1": 1,
"criteria_d_2": 0,
"criteria_d_3": 1
}
```

Figure 18: Prompt template used for Rubric-based ambient dimension evaluation